\documentclass[conference]{IEEEtran}
\usepackage{times}

\usepackage[numbers]{natbib}
\usepackage{multicol}
\usepackage[bookmarks=true]{hyperref}

\usepackage[english]{babel}

\usepackage[numbers]{natbib}

\usepackage{bibentry}

\usepackage{amsmath}

\usepackage{amssymb}

\usepackage{mathtools}

\usepackage{microtype}

\let\labelindent\relax
\usepackage[inline]{enumitem}

\usepackage{multicol}

\usepackage{booktabs}

\usepackage{subfig}

\usepackage[usenames,dvipsnames,table]{xcolor}

\usepackage{tcolorbox}

\usepackage{nag}

\usepackage{hyperxmp}

\usepackage{hyperref}
\definecolor{deeppink}{rgb}{1.0, 0.08, 0.58}
\hypersetup{
  colorlinks = true, 
  urlcolor   = deeppink, 
  linkcolor  = blue, 
  citecolor  = red 
}

\usepackage[acronym,toc]{glossaries}

\usepackage[export]{adjustbox}

\usepackage{tikz}

\usepackage[bottom,flushmargin]{footmisc}
\makeatletter  
\def\blfootnote{\gdef\@thefnmark{}\@footnotetext}
\makeatother

\usepackage{gensymb}

\usepackage{graphicx}

\usepackage{multirow}

\usepackage{makecell}

\newcolumntype{?}{!{\vrule width 1pt}}

\usepackage{pifont}

\usepackage{color, soul}

\newcommand{\pos}{\mathrm{pos}}
\newcommand{\fprate}{\afp}
\newcommand{\fnrate}{\afn}
\newcommand{\del}{\mathrm{d}}
\newcommand{\IG}{\mathrm{IG}}
\newcommand{\HH}{\mathrm{H}}
\newcommand{\afn}{\alpha_\mathrm{fn}}
\newcommand{\afp}{\alpha_\mathrm{fp}}
\newcommand{\fone}{F\textsubscript{1}}

\newcommand{\app}{App.}

\newcommand{\website}{https://siddancha.github.io/projects/active-velocity-estimation}

\begin{document}

\title{Active Velocity Estimation using Light Curtains\\via Self-Supervised Multi-Armed Bandits}

\author{  
  \IEEEauthorblockA{
    Siddharth Ancha\IEEEauthorrefmark{1}
    \hspace{2em}
    Gaurav Pathak\IEEEauthorrefmark{2}
    \hspace{2em}
    Ji Zhang\IEEEauthorrefmark{3}
    \hspace{2em}
    Srinivasa Narasimhan\IEEEauthorrefmark{3}
    \hspace{2em}
    David Held\IEEEauthorrefmark{3}\\
    \IEEEauthorrefmark{1}Massachusetts Institute of Technology, Cambridge, MA 02139, USA\\
    \IEEEauthorrefmark{2}Adobe, San Jose, CA 95110, USA\\
    \IEEEauthorrefmark{3}Carnegie Mellon University, Pittsburgh, PA 15213, USA
  }
  \\
  {\small Website: \href{\website}{\texttt{\website}}$^1$}
}

\maketitle

\begin{abstract}
  To navigate in an environment safely and autonomously, robots must accurately estimate where obstacles are and how they move.
Instead of using expensive traditional 3D sensors, we explore the use of a much cheaper, faster, and higher resolution alternative: \textit{programmable light curtains}. Light curtains are a controllable depth sensor that sense only along a surface that the user selects.
We adapt a probabilistic method based on particle filters and occupancy grids to explicitly estimate the position and velocity of 3D points in the scene using partial measurements made by light curtains.
The central challenge is to decide where to place the light curtain to accurately perform this task.
We propose multiple curtain placement strategies guided by maximizing information gain and verifying predicted object locations.
Then, we combine these strategies using an online learning framework. We propose a novel self-supervised reward function that evaluates the accuracy of current velocity estimates using future light curtain placements. We use a multi-armed bandit framework to intelligently switch between placement policies in real time, outperforming fixed policies.
We develop a full-stack navigation system that uses position and velocity estimates from light curtains for downstream tasks such as localization, mapping, path-planning, and obstacle avoidance.
This work paves the way for controllable light curtains to accurately, efficiently, and purposefully perceive and navigate complex and dynamic environments.\footnote{Please see our \href{\website}{project website} for (1) the appendix, (2) an overview video, (3) videos showing qualitative results of our method, and (4) source code.}
\end{abstract}

\blfootnote{\IEEEauthorrefmark{1}Corresponding author. E-mail: \texttt{sancha@mit.edu}.}
\blfootnote{\IEEEauthorrefmark{1}\IEEEauthorrefmark{2}This work was performed when SA and GP were affiliated with CMU.}

\IEEEpeerreviewmaketitle

\begin{figure*}[t!]
    \centering
    \subfloat[\scriptsize Light curtain working principle]{
        \includegraphics[trim=0 0 0 0,clip,width=0.475\textwidth]{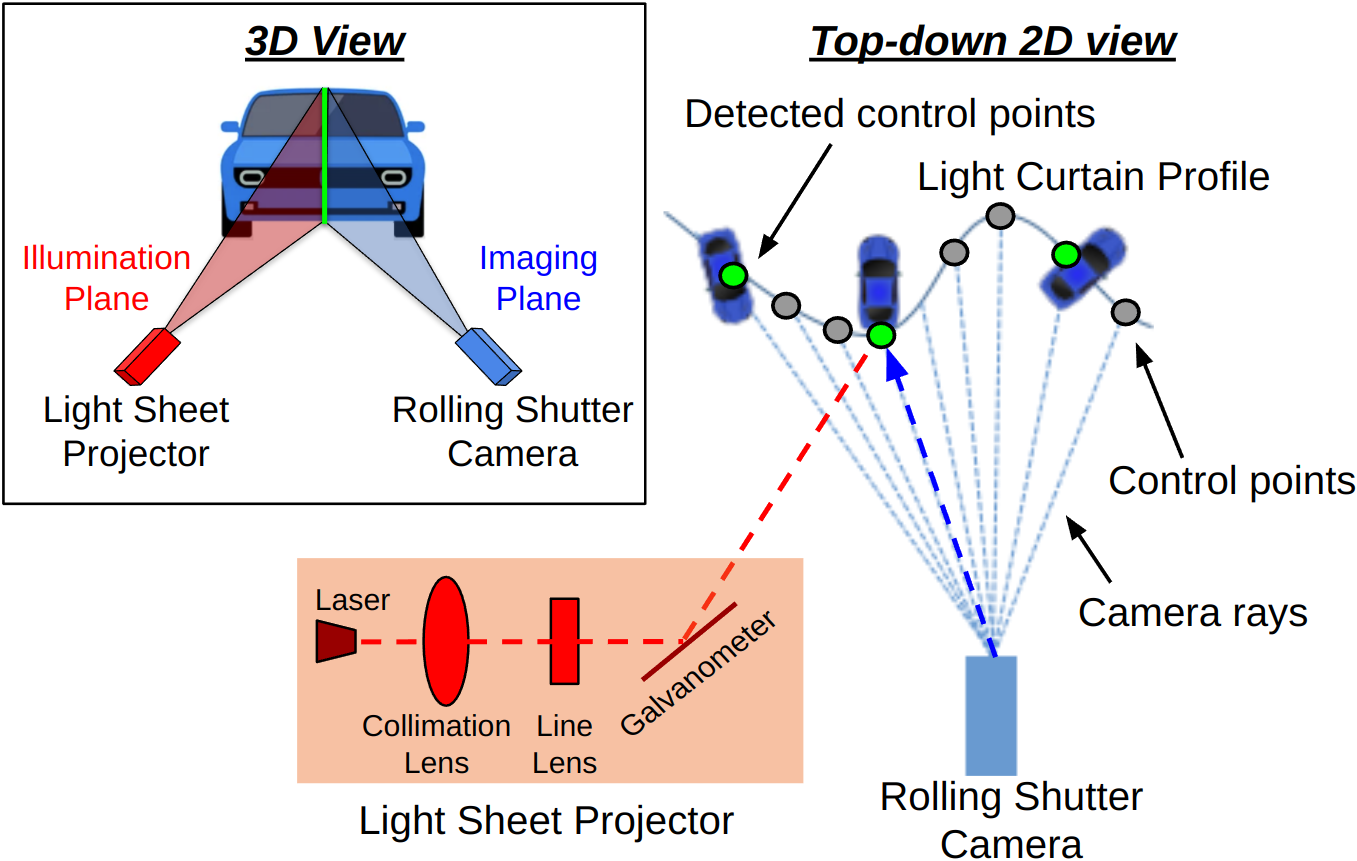}
    }\hspace{0.0em}
    \subfloat[\scriptsize Bayes filter with self-supervised reward]{
        \includegraphics[trim=0 0 0 0,clip,width=0.49\textwidth]{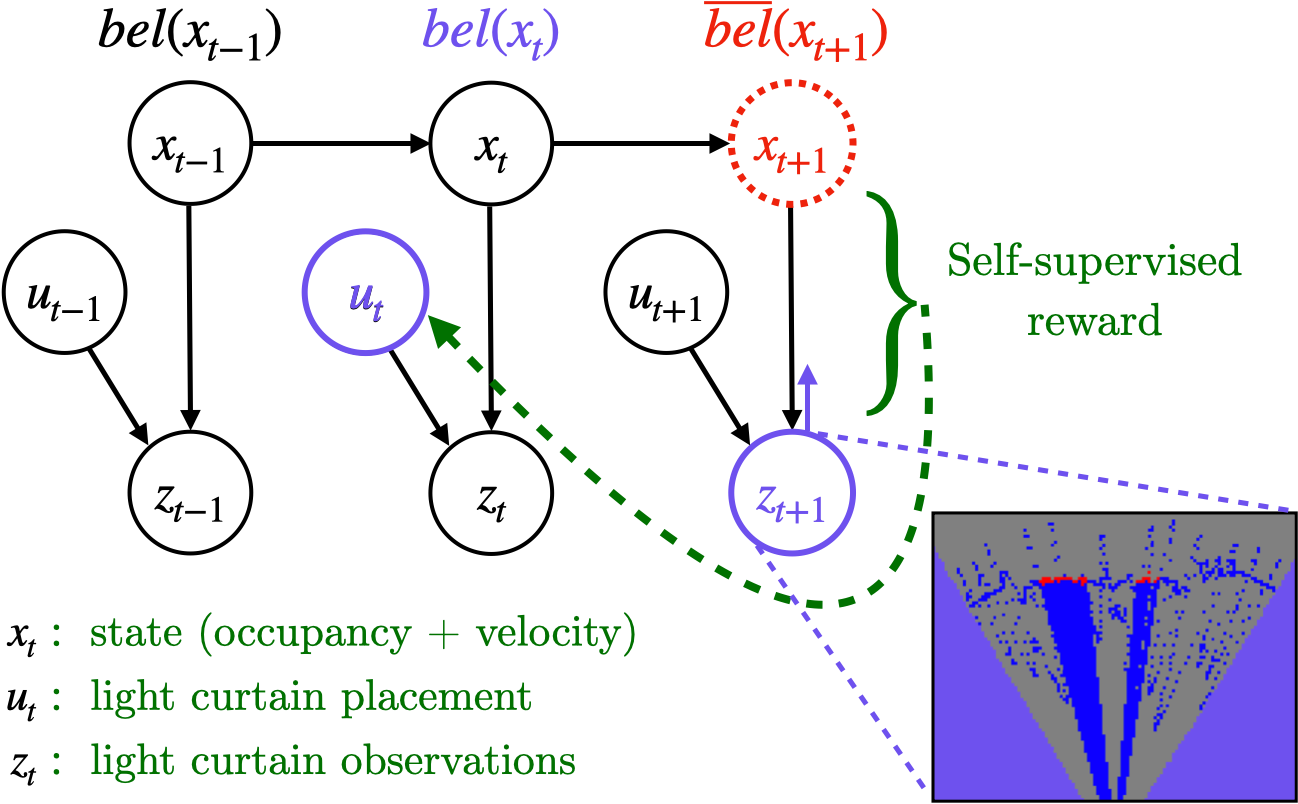}
    }
    \caption{
    \textbf{(a)}
    Illustration of programmable light curtains adapted from~\cite{ancha2020eccv}. An illumination plane (from the projector) and an imaging plane (of the camera) intersect to produce a light curtain. A controllable galvanometer mirror rotates synchronously with the camera's rolling shutter and images the points of intersection. See Sec.~\ref{sec:background-lc} for more details.
    \textbf{(b)}
    A \textit{Dynamic Bayes network}~\cite{thrun2002probabilistic} for controllable sensing. At timestep $t$, $x_t$ corresponds to the state of the world, $u_t$ corresponds to the action i.e. the location of light curtain placement,  $z_t$ corresponds to light curtain measurements, and $\overline{bel}(x_t)$ and $bel(x_t)$ are the inferred distributions over states before and after incorporating measurements $z_t$, respectively. This is a slightly modified graphical model for controllable sensing where actions $u_t$ don't affect state $x_t$ but directly affect observations $z_t$.}
    \label{fig:dbn}
  \end{figure*}
\section{Introduction}


Robots in the real world must navigate in the presence of moving objects like humans and vehicles whose motion is a priori unknown. This is a common challenge in many applications like autonomous driving, indoor and outdoor mobile robotics, and robot delivery. How should a robot sense and perceive such dynamic environments? How can it accurately estimate the motion of obstacles?

3D sensors such as LiDARs and depth cameras are conventionally used for robot navigation. However, LiDARs are typically expensive and low-resolution. Although cameras are cheaper and higher-resolution, depth estimates can be noisy and inaccurate. An alternative paradigm is \textit{active perception}~\cite{bajcsy1988active,bajcsy2018revisiting} where a controllable sensor is actively guided to focus on only the relevant parts of the environment. Programmable light curtains~\cite{wang2018programmable,bartels2019agile,ancha2020eccv,raaj2021cvpr,ancha2021rss} are a recently invented, lightweight 3D sensor that detects points intersecting any user-specified 2D surface (``curtain"). Light curtains combine the best of passive cameras (low cost, high resolution, and high speed) and LiDARs (accurate depth estimation along the curtain, robustness to bright lighting and scattered media like fog/smoke~\cite{wang2018programmable}).
Compared to widely used commercial LiDARs like the Ouster OS1-128~\cite{ousteros1}, a lab-built light curtain prototype is relatively inexpensive (\$1,000 v.s. $\sim$\$20,000), higher vertical resolution (1280 rows/$0.07^\circ$ v.s. 128 rows/$0.35^\circ$) and faster (45-60 Hz v.s. 10-20 Hz).
See \app~\ref{app:lc-sensor-benefits} for benefits of light curtains over conventional depth sensors.
Because programmable light curtains are an active sensor, 
realizing these benefits requires actively deciding where to place the curtain at each timestep; this is the principal algorithmic challenge posed by programmable light curtains.

Previously, light curtains have been used for object detection~\cite{ancha2020eccv}, depth estimation~\cite{raaj2021cvpr}, and estimating safety regions~\cite{ancha2021rss}. However, light curtains have not been used to explicitly estimate velocities of dynamic objects. Velocity estimation is crucial for many tasks in robotics such as trajectory forecasting, obstacle avoidance, motion planning, and dynamic object removal for SLAM~\cite{thrun2002probabilistic}.

The focus of this paper is to develop light curtain placement strategies that improve velocity estimates.
We use dynamic occupancy grids~\cite{danescu2011modeling} to estimate velocities and occupancies from points detected by light curtains without requiring point cloud segmentation or explicit data association across frames.
First, we extend light curtain placement strategies from previous works~\cite{ancha2020eccv,ancha2021rss} to integrate dynamic occupancy grids. Then, we propose a novel method to switch between multiple light curtain placement strategies using a multi-armed bandits approach.
The feedback for the multi-armed bandits is obtained using a novel \textit{self-supervised} reward function that evaluates the current estimates of occupancy and velocity using future light curtain placements, without requiring ground truth or additional sensors. We obtain this supervision by reusing intermediate quantities computed during recursive Bayes estimation of dynamic occupancy grids; thus the self-supervised rewards do not require extra light curtain placements or additional computations.
We evaluate our approach on challenging simulated and real-world environments with complex and fast object motion. We integrate our method into a full-stack navigation pipeline and show that the multi-armed bandits approach is able to outperform each individual strategy.

Our contributions include:
\begin{enumerate}[leftmargin=*]
    \item We re-derive the dynamic occupancy grid method~\cite{danescu2011modeling} using a more rigorous mathematical analysis grounded in Bayesian filtering~\cite{thrun2002probabilistic} (Sec.~\ref{sec:dynamic-occupancy-grids}, \app~\ref{app:dynamic-occupancy-grids}). 
    \item We design curtain placement strategies for dynamic occupancy grids to verify predicted object locations (Sec.~\ref{sec:max-depth-prob}) and maximize information gain in hybrid discrete-continuous spaces (Sec.~\ref{sec:max-info-gain}, \app~\ref{app:max-info-gain}).
    \item We propose a novel \textit{self-supervised} reward function that evaluates current velocity estimates using future light curtain placements without requiring additional supervision.  Using the self-supervised reward, we learn to combine multiple curtain placement policies using a multi-armed bandit framework (Sec.~\ref{sec:mab}).
    \item We evaluate this approach in simulated and real-world environments with fast-moving obstacles and demonstrate that it outperforms individual placement strategies (Sec.~\ref{sec:experiments}).
    \item We develop an efficient and parallelized pipeline where light curtain sensing, grid estimation and computing curtain placement are tightly coupled and continuously interact with each other at $\sim$45 Hz (Sec.~\ref{sec:parallelized-pipeline}, Fig.~\ref{fig:parallelized-pipeline}, \app~\ref{app:extra-grid}).
    \item We integrate our method into a full-stack navigation pipeline that uses position and velocity estimates to perform localization, mapping and obstacle avoidance in real-world dynamic environments (\app~\ref{app:full-stack-navigation}).
\end{enumerate}


\section{Background}
\label{sec:background}

\subsection{Light curtain working principle}
\label{sec:background-lc}

Programmable \textit{light curtains}~\cite{wang2018programmable,bartels2019agile,ancha2020eccv,raaj2021cvpr,ancha2021rss} are a recently developed controllable depth sensor that image any user-specified vertically-ruled 2D surface in the environment.
The device contains two main components: a rolling-shutter camera and a rotating light sheet laser (see illustration in Fig.~\ref{fig:dbn}a).
The camera activates one pixel column at a time, from left to right, via the rolling shutter. We refer to the top-down projection of the imaging plane corresponding to each pixel column as a ``camera ray'' (shown in Fig.~\ref{fig:dbn}a). The shape of the light curtain is entirely specified by a 2D \textit{control point} selected on each camera ray (shown as gray and green circles). The set of control points forms the input to the light curtain device. The laser is vertically aligned and synchronized with the camera's rolling shutter. A controllable galvo-mirror rotates the light sheet to point it at the control point corresponding to the currently active pixel column. Triangulated 3D scene points that both (1) intersect the laser light sheet and (2) are visible in the currently active pixel column, get detected by the device. 
If there exists an object in the environment at the surface of this intersection, then the point will have a large intensity in the camera reading; otherwise it will not; thus the device outputs the subset of control points (shown as green circles in Fig.~\ref{fig:dbn}a) that correspond to 3D object surfaces. 
Importantly, light curtains form a partial observation on the scene, since only control points can be detected.
Please see \cite{bartels2019agile,wang2018programmable} for further details on the mechanism behind a programmable light curtain.

\subsection{Bayes filtering}
\label{sec:background-bf}

This section provides a brief background on Bayes filtering and introduces notation used throughout the paper. A \textit{dynamic Bayes filter}~\cite{thrun2002probabilistic}, also known as a \textit{hidden Markov model} or a \textit{state space model} is represented by a probabilistic graphical model shown in Fig.~\ref{fig:dbn}b. The state of the world at timestep $t$ is denoted by $x_t$ (in our case, $x_t$ is the occupancy and velocity of a set of cells arranged in a 2D grid from the top-down view; more details in \app~\ref{app:mathematical-framework}).
The control actions are denoted by $u_t$ (the locations where the light curtain is placed). Observations obtained from the sensor are denoted by $z_t$. Fig.~\ref{fig:dbn}b is a slight modification of the standard model for the task of active perception, where actions don't affect the state of the world $x_t$ but directly affect the observations $z_t$.

The goal is to infer at each timestep $t$ the posterior distribution (a.k.a ``belief") $bel(x_t) = P(x_t \mid u_{1:t}, z_{1:t})$ over the current state $x_t$ from the sequence of sensor observations $z_{1:t}$ and the known sequence of actions $u_{1:t}$.
This is computed using \textit{recursive Bayesian estimation}~\cite{thrun2002probabilistic} that alternates between two steps.
\begin{align}
\overline{bel}(x_t) &= \int_{x_{t-1}}~bel(x_{t-1})~P(x_t \mid x_{t-1})~\del x_{t-1}
\label{eqn:generic-motion-update}\\
bel(x_t) &\propto P(z_t \mid x_t, u_t)~\overline{bel}(x_t)
\label{eqn:generic-measurement-update}
\end{align}
First, the \textit{motion update} step computes an intermediate prior belief $\overline{bel}(x_t)$ by applying a motion model $P(x_t \mid x_{t-1})$ that encodes the dynamics of the environment. Then, the \textit{measurement update} step computes the updated posterior belief $bel(x_t)$ by incorporating sensor observations from the current timestep. To make this paper self-contained, we provide a detailed mathematical derivation of these steps in \app~\ref{app:bayes-filtering}.

\begin{figure*}[t!]
    \centering
    \subfloat[\scriptsize Dynamic occupancy grid: each cell contains occupancy and velocity distributions]{
        \includegraphics[trim=0 0 0 0,clip,width=0.44\textwidth]{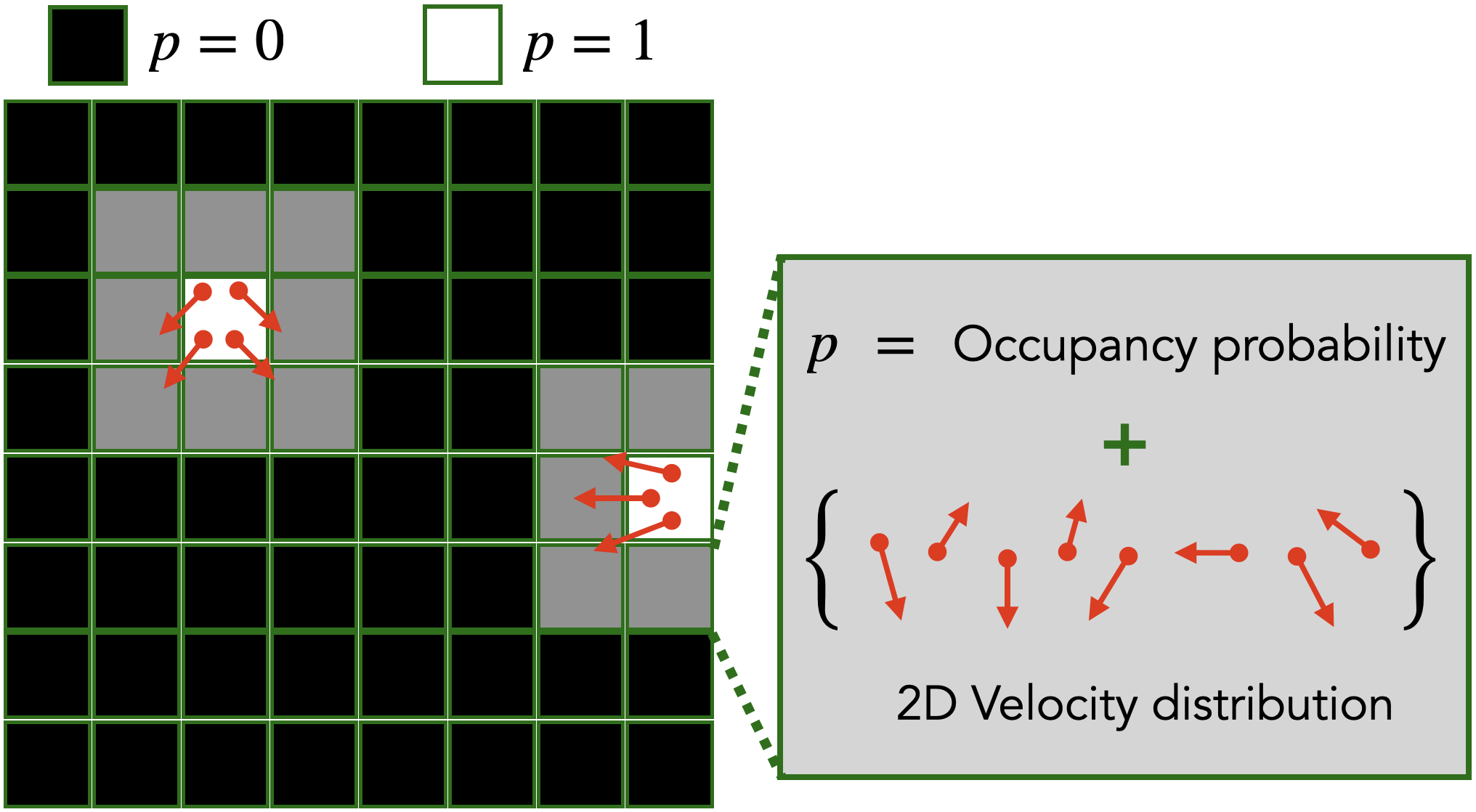}
    }
    \hspace{0.5em}
    \subfloat[\scriptsize Raycasting to compute freespace]{
        \includegraphics[trim=0 -50 0 0,clip,width=0.245\textwidth]{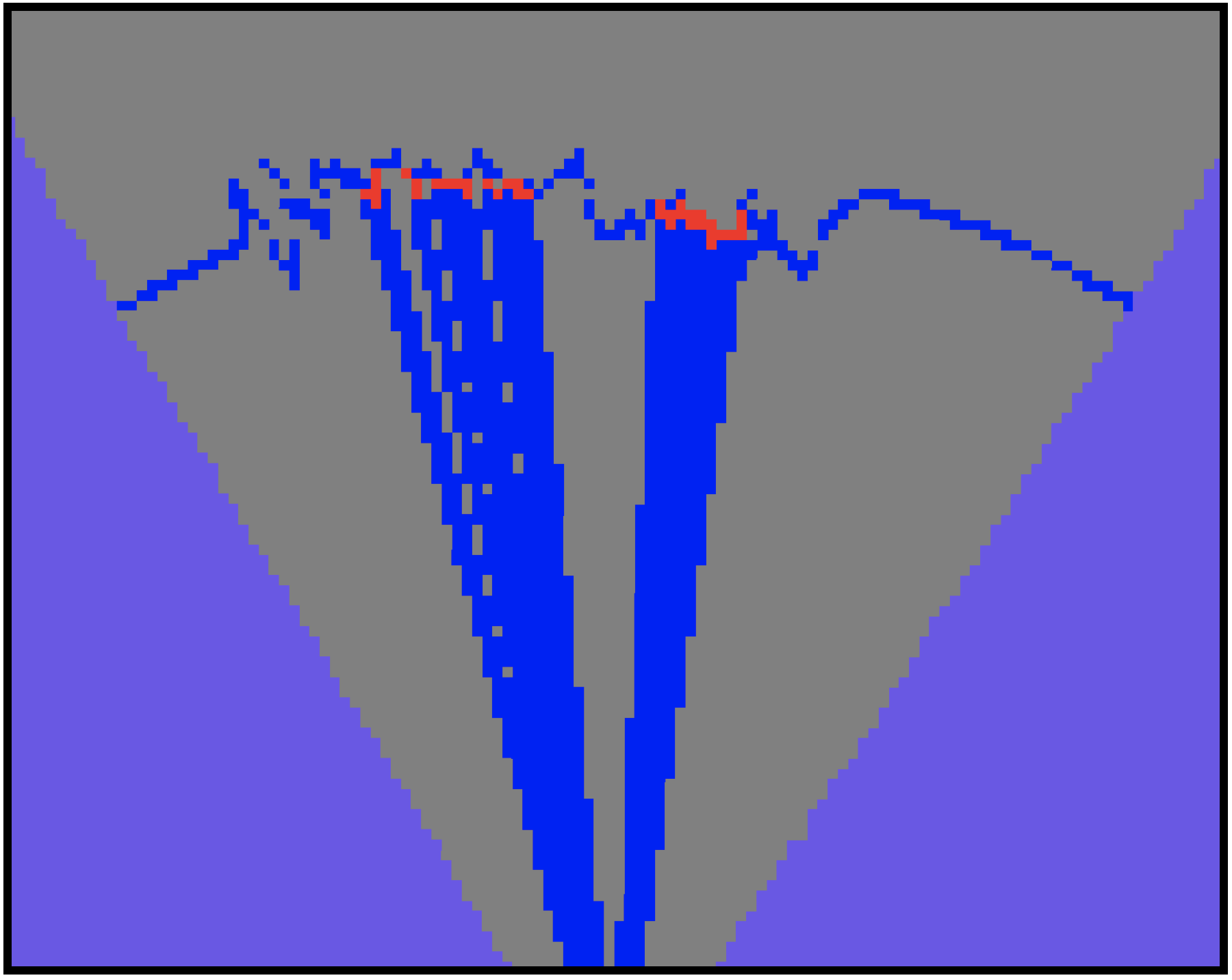}
    }
    \hspace{0.5em}
    \subfloat[\scriptsize Raymarching to compute depth probs.]{
        \includegraphics[trim=0 0 0 0,clip,width=0.22\textwidth]{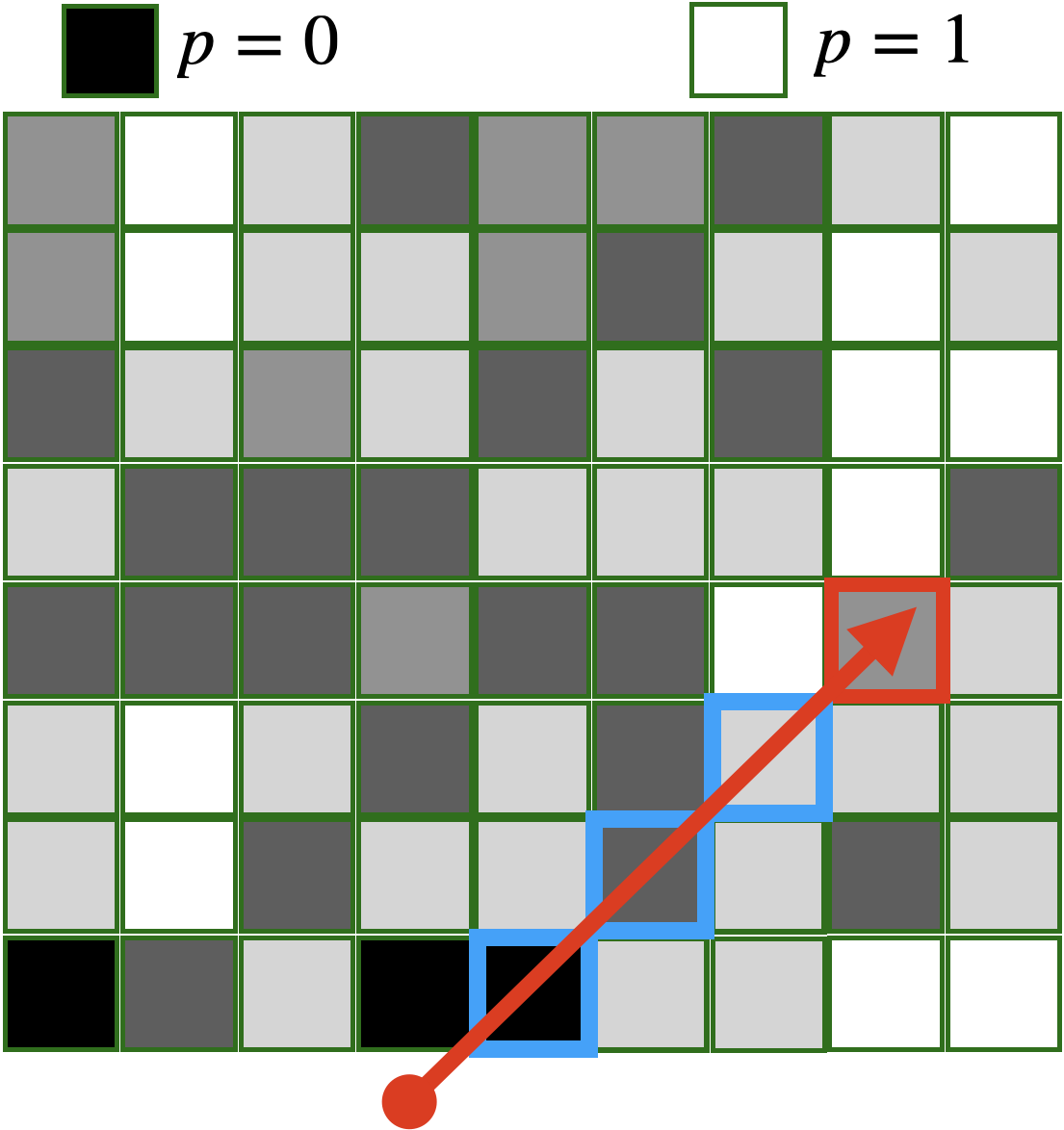}
    }
    \caption{
        \textbf{(a)} \textit{Dynamic occupancy grid.} The 2D grid represents the top-down view. Like conventional occupancy grids~\cite{elfes1989using,thrun2002probabilistic}, each cell contains an occupancy probability $p \in [0, 1]$. In addition, each cell also contains a set of weighted \textit{particles} where each particles stores a single 2D velocity. The set of particles together represents a probability distribution of that cell's velocity.
        \textbf{(b)} \textit{Ray-casting} to light curtain detections to extract freespace information. Red cells contain detected points and are marked occupied. Blue cells are freespace; they either lie undetected on the light curtain or lie on rays cast from the sensor to the red cells. Gray denotes unknown occupancy. Purple cells are outside the light curtain's field of view.
        \textbf{(c)} \textit{Ray-marching} to compute the depth probabilities of cells along a camera ray. The depth probability of the red cell is the product of the probability that the red cell is occupied and the probabilities of each blue cell being unoccupied.
    }
    \label{fig:dog}
\end{figure*}

\section{Related Work}

\subsection{Active perception and light curtains}

Active perception involves actively controlling a sensor such as camera parameters~\cite{bajcsy1988active}, moving a camera to look around occlusions~\cite{cheng2018reinforcement}, and next-best view planning~\cite{connolly1985determination} for object instance classification~\cite{wu20153d,doumanoglou2016recovering,denzler2002information,scott2003view} and 3D reconstruction~\cite{isler2016information,kriegel2015efficient,vasquez2014volumetric,daudelin2017adaptable}.
Programmable light curtains~\cite{wang2018programmable,bartels2019agile,chan2022holocurtains} are a controllable depth sensor that have been used for active perception tasks such as active object detection~\cite{ancha2020eccv}, active depth estimation~\cite{raaj2021cvpr}, and actively estimating safety regions~\cite{ancha2021rss}.
However, most prior light curtain work has only focused on estimating object positions. They either place curtains with fixed scan patterns~\cite{wang2018programmable,chan2022holocurtains} or adaptive curtains for static scenes without taking object motion into account~\cite{bartels2019agile,ancha2020eccv,raaj2021cvpr}. \citet{ancha2021rss} track safety regions by learning to forecast future locations; this could be interpreted as \textit{implicit} velocity estimation. However, we are the first to explicitly estimate obstacle velocities which can be used for other downstream tasks like trajectory forecasting, obstacle avoidance and motion planning. Furthermore, we combine multiple adaptive strategies like random curtains~\cite{bartels2019agile,ancha2021rss}, maximizing information gain~\cite{ancha2020eccv}, and verifying predicted object locations~\cite{ancha2021rss} using a novel multi-armed bandit framework for estimating both object positions and velocities.

\subsection{Velocity estimation from point clouds}
Prior works on estimating \textit{scene flow}~\cite{vedula1999three,wang2018deep,liu2019flownet3d,gu2019hplflownet,teed2021raft} compute correspondences between point clouds acquired at consecutive timesteps; velocities can then be extracted from these correspondences. Furthermore, self-supervised approaches~\cite{mittal2020just,wu2020pointpwc,kittenplon2021flowstep3d,li2021self,baur2021slim,rangarajan2022alignment} can learn to estimate scene flow without requiring ground truth annotations. However, these methods are designed to compute flow between \textit{complete} scans of the environment, such as those obtained from a LiDAR sensor, where correspondences exist for most points. In contrast, a single light curtain measurement is a \textit{partial} point cloud -- a subset of visible points that intersect the curtain. Depending on where they are placed, consecutive light curtains may not contain any correspondences at all. Therefore, scene flow methods are not suited for point clouds acquired by light curtains.

Another approach is to first segment the point cloud into a collection of separate objects~\cite{held2016probabilistic,douillard2011segmentation,klasing2008clustering,teichman2011towards,zhao2022divide,hu2020learning}, track each object, and finally register each object's segmented point cloud across frames using either optimization-based~\cite{besl1992method,rusu2009fast,held2013precision,zhou2016fast,yang2020teaser,makadia2006fully}, probabilistic~\cite{held2014combining,held2016robust,hahnel2002probabilistic,agamennoni2016point}, or learning-based~\cite{wang2019deep,wang2019prnet,aoki2019pointnetlk,choy2020deep} methods. However, errors in point cloud segmentation can lead to incorrect velocity estimates. Instead, our method uses particle-based occupancy grids and avoids the need to perform either segmentation or explicit data association across frames.

\subsection{Self-tuning Bayes filters}

Prior works have used \textit{innovation} i.e. the difference between predicted and observed measurements of a Kalman filter, to ``self-tune'' model parameters without needing ground truth annotations. Earlier works use an autoregressive moving average innovation model (ARMA)~\cite{hagander1977self,moir1984optimal,fung1983dynamic,deng2008self,zi2007self}. More recent works use the \textit{normalized innovation squared (NIS)} metric to optimize Kalman filter noise models using downhill simplex methods~\cite{powell2002automated}, Bayesian optimization~\cite{chen2018weak,chen2019kalman}, and evolutionary algorithms~\cite{oshman2000optimal,cai2019towards}. Our self-supervised metric is inspired by Kalman filter innovation,
but is used to select a sensor control strategy at each timestep using multi-armed bandits rather than tuning noise models.

\section{Dynamic occupancy grids}
\label{sec:dynamic-occupancy-grids}

We now describe how we apply
\textit{dynamic occupancy grids}~\cite{danescu2011modeling} for velocity estimation with light curtains. A dynamic occupancy grid is a Bayes filter that combines two conventional representations in robotics: occupancy grids and particle filters.
Occupancy grids~\cite{elfes1989using,thrun2002probabilistic} are a standard tool for mapping the location of static objects in the environment from the 2D top-down view. Each cell in the grid contains an \textit{occupancy probability} $p\in [0, 1]$, denoting the probability of the cell being occupied by an object.
\textit{Dynamic} occupancy grids~\cite{danescu2011modeling} are an extension of classical occupancy grids (see Fig.~\ref{fig:dog}a). Each cell in the grid contains both the occupancy probability $p$ as well as a probability distribution over 2D velocities. The velocity distribution is represented by a set of weighted particles, where each particles stores a single 2D velocity. The set of weighted particles approximates the true velocity distribution.

While Danescu et. al.~\cite{danescu2011modeling} showed that dynamic occupancy grids can accurately estimate occupancies and velocities, the precise role of particles and what they represent remained unclear. Particles were described as representing a cell's velocity distribution; however, the movement of particles from one cell to another is somewhat inconsistent with this interpretation.  Elsewhere, particles are described 
as being ``physical building blocks of the world'', i.e. parts of objects that can move; however, under this interpretation, it is unclear what distribution a set of particles is supposed to represent, since each particle represents a different part of an object.
Furthermore, the particles were not only used to represent velocities, but their count inside a cell was proportional to the occupancy probability. In this work, we re-derive dynamic occupancy grids using a more rigorous mathematical analysis found in \app~\ref{app:dynamic-occupancy-grids}, in which we explicitly state the assumptions made and provide a precise, mathematically rigorous interpretation of particles. 


\textbf{Motion and measurement updates:} In the motion update step, particles are resampled from each cell in the grid and moved to another cell based on their velocities and the motion model. We assume access to a  depth sensor (e.g. light curtains, LiDAR, depth cameras) that measures depth but does not directly measure velocity. In the measurement update step, the sensor provides (noisy) observations of occupancy for a subset of un-occluded cells in the grid. These observations are used to update the occupancy probabilities; velocities are inferred indirectly in the motion update step that are consistent with observed occupancies. This method is able to estimate velocities from depth measurements alone without requiring explicit data association across frames.

\textbf{Raycasting to extract freespace information:} As explained in Section~\ref{sec:background-lc}, 
a light curtain only returns whether there is a 3D object surface at the location of the control points where the camera rays and the laser sheets intersect; no depth information is returned for other locations in the environment. 
Fig.~\ref{fig:dog}b shows an observation grid from a light curtain placement where cells directly measured to be occupied are shown in red and free cells are shown in blue. From this figure, we see that all voxels in between the light curtain source and a detected point must be unoccupied. Since 3D points were detected in the occupied cells, light must have traveled along these rays without obstruction; we mark cells along these rays to be free (shown in blue in Fig.~\ref{fig:dog}b). 
To take advantage of this information, we cast rays using an efficient voxel traversal algorithm~\cite{amanatides1987fast,hu2020you} from the sensor to occupied cells (shown in red). More details can be found in \app~\ref{app:measurement-update-step}. Thus by exploiting visibility constraints, we are able to extract more information from the light curtain.

\begin{figure*}[ht!]
    \centering
    (a)
    {
        \includegraphics[trim=0 0 0 0,clip,width=0.27\textwidth,frame=0pt]{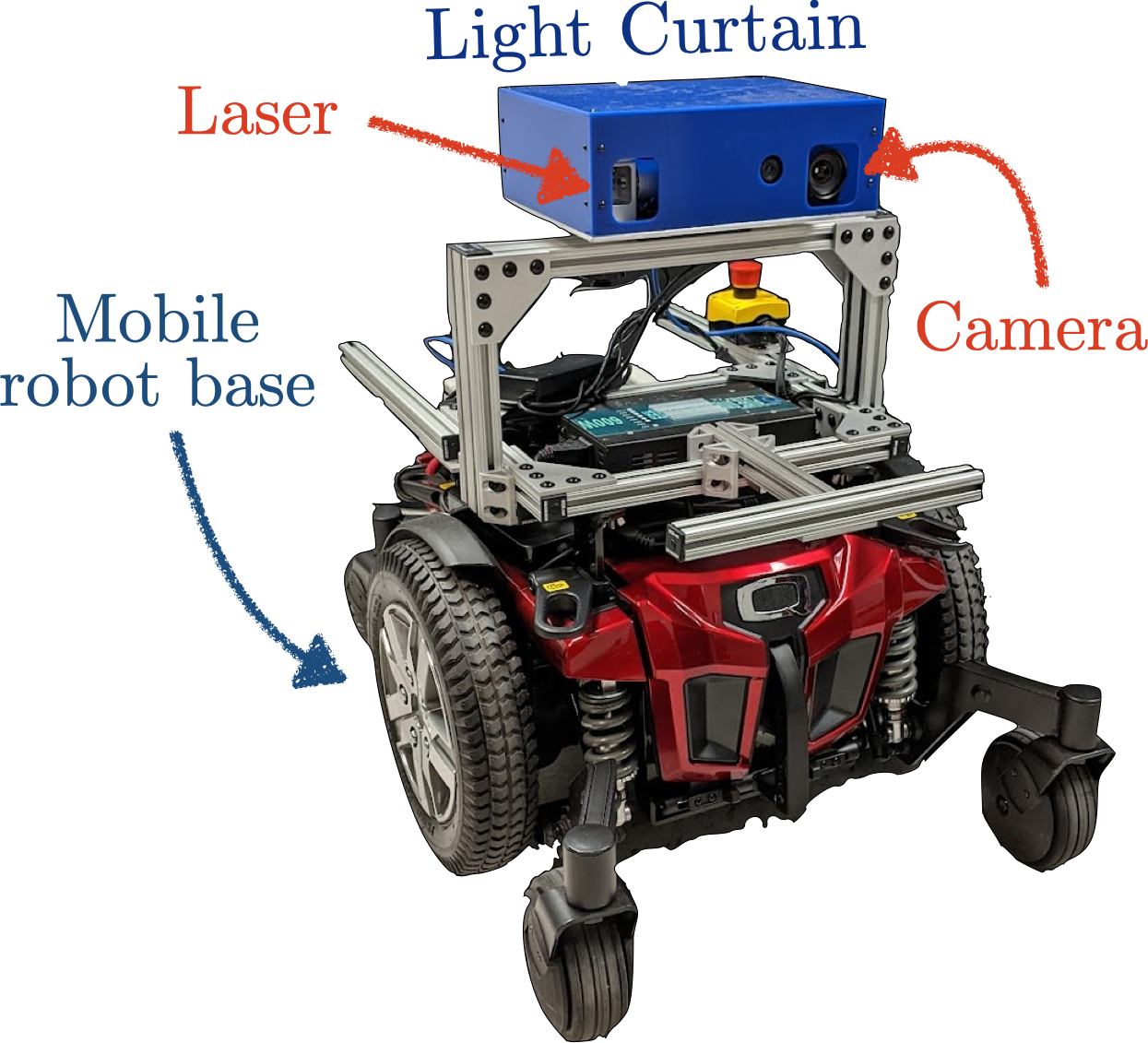}
    }
    (b)
    {
        \includegraphics[trim=0 0 0 0,clip,width=0.3\textwidth]{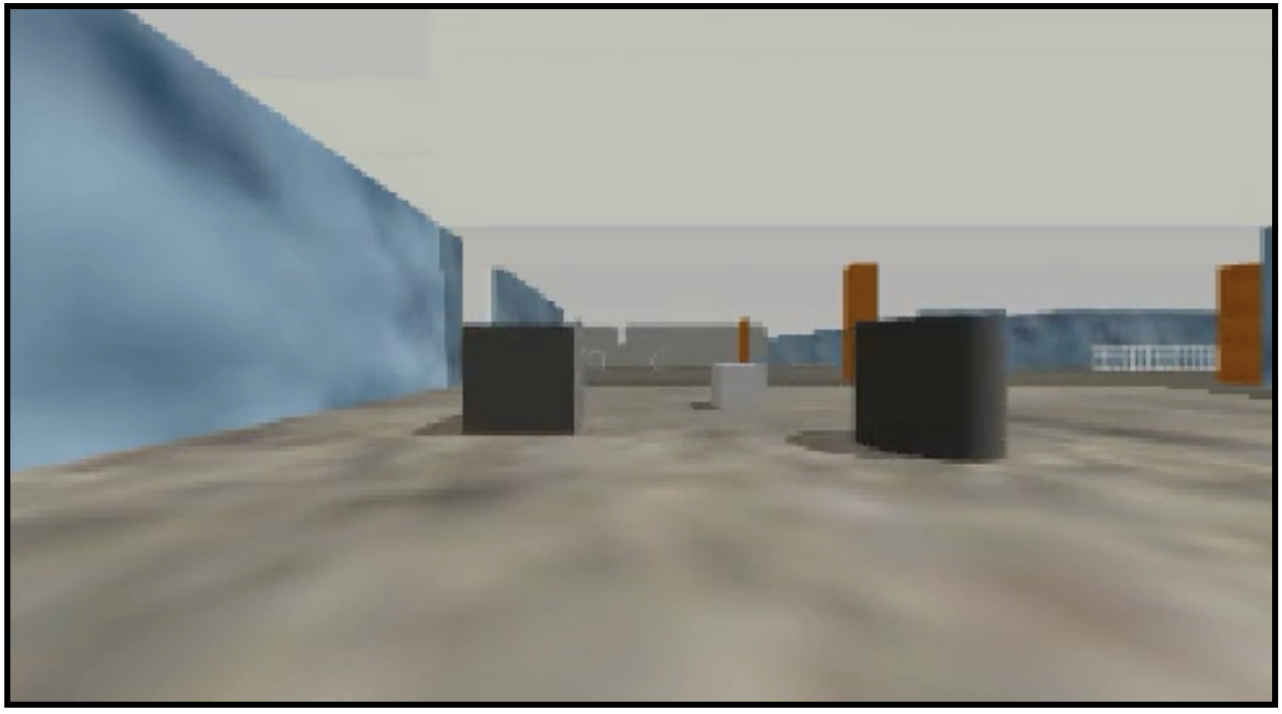}
    }
    (c)
    {
        \adjustbox {padding=-3.5pt 0pt -3.5pt 0pt,margin=0pt 0pt 0pt 0pt,frame=0.5pt}
        {
            \includegraphics[trim=0 0 0 0,clip,width=0.25\textwidth]{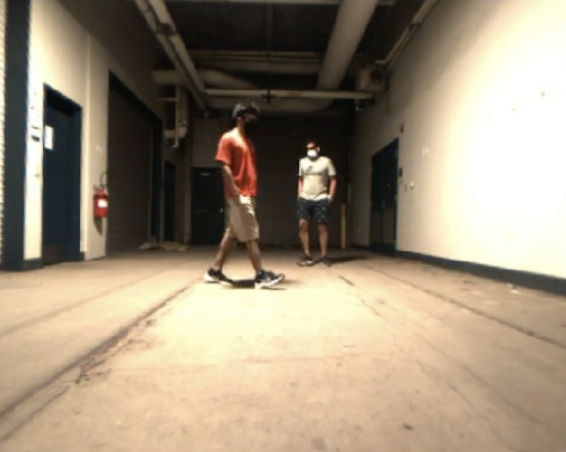}
        }
    }
    \caption{
        \textbf{(a)}
        \textit{Mobile light curtain robot platform:} A light curtain device (in blue) is mounted on top of a mobile robot. We use this setup to perform real-world experiments.
        \textbf{(b)}
        \textit{Simulated environment:} consists of differently shaped objects (cuboids and cylinders) moving in (1) linear oscillatory/harmonic motion along various directions, (2) curved sinusoidal motion, and (3) random Brownian motion.
        \textbf{(c)}
        \textit{Real-world environment:} consists of two pedestrians walking in front of the sensor in multiple directions, at different speeds and in complex trajectories.}
    \label{fig:wheelchair-envs}
\end{figure*}

\section{Curtain placement strategies}
\label{sec:intelligent-curtain-strategies}

\newcommand{\Pd}{P^\mathrm{D}_t}
\newcommand{\Pv}{P^\mathrm{V}_t}

Using dynamic occupancy grids and Bayesian filtering, 
we have a method to infer occupancies and velocities explicitly from light curtain measurements (details in \app~\ref{app:dynamic-occupancy-grids}). The main challenge that we address in this paper is to compute the best curtain placement from the dynamic occupancy grid i.e. from the current estimates of occupancy and velocity. The measurements from the placed curtain will be input back to upgrade the grid, closing the loop.

In order to compute the best curtain placement, we must first predict the occupancy when the next light curtain will be placed. To do so, we forecast the current dynamic occupancy grid, using the currently estimated velocities, to the next timestep via the motion update step (Eqn.~\ref{eqn:generic-motion-update}, Eqn.~\ref{eqn:dog-motion-update} in App.~\ref{app:motion-update-step}).
In this section, we propose various curtain placement strategies computed from the forecasted grid. In Sec.~\ref{sec:mab}, we will propose a novel method to combine them and outperform each individual strategy.

\subsection{Maximizing depth probability}
\label{sec:max-depth-prob}

\textbf{Strategy 1: Depth Probability:} Following \citet{ancha2021rss}, Strategy 1 places curtains at the highest probability object locations.
This strategy is motivated by the fact that a light curtain only senses visible object surfaces when it intersects them. Therefore, this approach can be used to verify whether objects are indeed located at the forecasted object locations.

Since occupancy grids are probabilistic, this strategy places curtains at locations of highest ``depth probability'', which is the probability that a control point at a given cell would return a depth reading. The depth probability of a cell is the probability that the cell is occupied, and all occluding cells (lying on the ray starting from the sensor and ending at the target cell) are free (see Fig.~\ref{fig:dog}c).
We borrow the idea of ``ray marching'' from the literature on volumetric rendering~\cite{tulsiani2017multi,mildenhall2020nerf} to compute depth probabilities efficiently; see \app~\ref{app:max-depth-prob} for more details on the algorithm and computational complexity. For each camera ray, we place the curtain on the cell with the maximum depth probability.

\subsection{Maximizing information gain}
\label{sec:max-info-gain}

Another placement strategy that was found useful in previous work on 3D object detection~\cite{ancha2020eccv} was to place curtains at the regions of highest ``uncertainty''. This is based on the principle of maximizing \textit{information gain} for active sensing.

Recall the dynamic Bayes network in Fig.~\ref{fig:dbn}b. Given a forecasted prior belief $P(x_t) = \overline{bel}(x_t)$, the information gain framework prescribes that the action $u_t$ should be taken that maximizes the information gain $\IG(x_t, z_t \mid u_t)$ between the state $x_t$ and the observations $z_t$ when using $u_t$. Information gain, which is a well-studied quantity in information theory, is the expected reduction in entropy (i.e. uncertainty) before and after sensing:
$\HH(P(x_t)) - \mathbb{E}_{z_t \mid u_t} \big[ \HH(P(x_t \mid z_t, u_t)) \big]$.



While information gain for conventional occupancy grids is straightforward to derive~\cite{ancha2020eccv}, it is not so for the case of dynamic occupancy grids. This is because the underlying state space of dynamic occupancy grids is a `mixture' of discrete and continuous spaces -- a cell can either be unoccupied or occupied with a continuous velocity. Unfortunately, the entropy of such mixed discrete-continuous spaces is not well-defined~\cite{gao2017estimating}.
We overcome this problem using a more general definition of information gain based on the ``Radon–Nikodym'' derivative~\cite{gao2017estimating} that doesn't require explicitly calculating the entropy. In \app~\ref{app:max-info-gain}, we show that the formula for information gain for dynamic occupancy grids (under certain assumptions) turns out to equal the occupancy uncertainty, described next.

\textbf{Strategy 2: Occupancy Uncertainty}:
Let $\omega^i_t$ be the occupancy probability estimated for the $i$-th cell at the $t$-th timestep. Then, the information gain is the sum of binary cross entropies $\HH_\text{occ}(\omega^i_t) = -\omega^i_t~\log_2\omega^i_t -(1-\omega^i_t)~\log_2(1-\omega^i_t)$ of the cells that the curtain lies on.
Intuitively, since measurements from a depth sensor only provide information about occupancy and not velocity, the overall information gain is equal to the total occupancy uncertainty. A similar information gain computation was used in \citet{ancha2020eccv} for static occupancy grids; in \app~\ref{app:max-info-gain} we prove that the formula for information gain is the same as total occupancy uncertainty even for the more complex case of mixed discrete-continuous distributions. Strategy 2 places a curtain that maximizes the occupancy uncertainty.

\textbf{Strategy 3: Velocity Uncertainty}: Each cell also contains a velocity distribution $V^i_t = \{(v^{i, m}_t, p^{i, m}_t) \mid 1 \leq m \leq M \}$ represented by a set of $M$ weighted particles with velocities $v^{i, m}_t$ and weights $p^{i, m}_t$ that sum to $1$. In this strategy, we maximize the sum of velocity entropies.
The discrete set of particles is used to approximate what is inherently a continuous velocity distribution.
Therefore, we must compute the \textit{differential} entropy of the continuous velocity distribution by first estimating its probability density function.
We fit a multivariate Gaussian distribution to the set of weighted particles with mean $\mu = \sum_{m=1}^M p^{i, m}_t v^{i, m}_t$ and covariance matrix $\Sigma = \sum_{m=1}^M p^{i, m}_t (v^{i, m}_t - \mu) (v^{i, m}_t - \mu)^T$. Then, we compute the differential entropy of the fitted Gaussian: $\HH_\text{vel}(V^i_t) = \frac{1}{2} \log \det (2\pi e \Sigma)$. One could alternatively use other families of continuous distributions, such as kernel density estimators. Finally, we place a curtain that maximizes the sum of velocity entropies $\HH_\mathrm{vel}$ of the cells the curtain lies on.


\textbf{Strategy 4: Combined  Uncertainty}: In this strategy, we maximize a weighted combination of occupancy and velocity entropies: $\HH_\mathrm{cmb}(\omega^i_t, V^i_t) = \HH_\mathrm{occ}(\omega^i_t) + \omega^i_t~\HH_\mathrm{vel}(V^i_t)$. The velocity uncertainty is weighted by the occupancy probability. This captures the notion that if the occupancy probability is very low, then the overall uncertainty should also be low even if the velocity uncertainty is high, because the velocity uncertainty is not relevant if the cell is unoccupied. This is a heuristic curtain placement policy that performs well in practice.


\begin{figure*}[h!]
    \centering
    \includegraphics[trim=0 0 0 0,clip,width=0.9\textwidth]{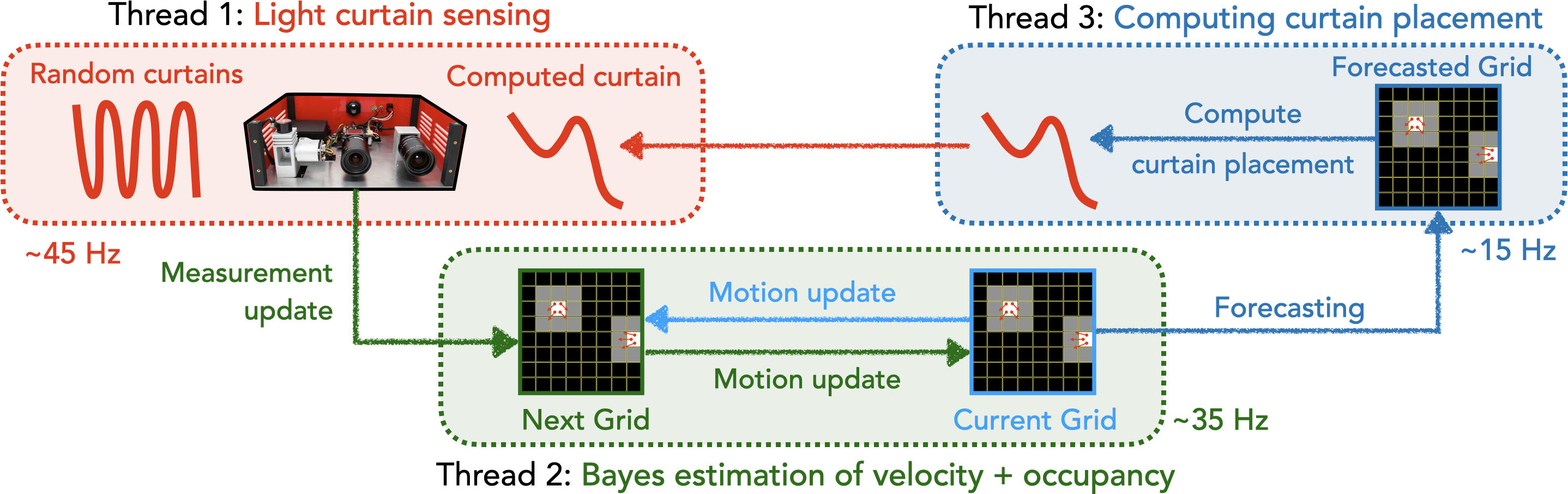}
    \caption{\textit{Implementation of our method as a parallelized pipeline.} Our methods contains three components: (1) light curtain sensing, (2) Bayes estimation of dynamic occupancy grids, and (3) computing curtain placement. Each process can be run in parallel in a separate thread at its own independent speed. The three processes are tightly coupled in a closed loop using three grids as shared memory. Our implementation ensures that information flows between the threads safely and continuously.}
    \label{fig:parallelized-pipeline}
\end{figure*}

\section{Self-supervised multi-armed bandits}
\label{sec:mab}


Can we combine the various curtain placement strategies developed in Sec.~\ref{sec:intelligent-curtain-strategies} to improve performance? In this section, we develop a multi-armed bandit method to do so enabled by a novel \textit{self-supervised} reward function. 

\subsection{Multi-armed bandit framework}
A multi-armed bandit~\cite{sutton2018reinforcement} is an online learning framework consisting of a set of actions or ``arms'', where each action is associated with an unknown reward function.
The agent only observe samples from the reward distribution when it takes that action. The goal is to maximize the cumulative reward over time. The agent maintains a running average of the rewards for each action, called \textit{Q-values}.
We use $\epsilon$-greedy multi-armed bandits~\cite{sutton2018reinforcement}, that trades-off exploration with exploitation. With probability $\epsilon$, the bandit performs exploration and chooses an action at random. With probability $1-\epsilon$, it performs exploitation and chooses the action that has the highest Q-value. We use multi-armed bandits to intelligently switch between the four curtain placement strategies at test time.

\subsection{Self-supervised rewards}
\label{sec:ss-rewards}

The bandit framework requires a reward function to evaluate actions. Our eventual goal is to accurately estimate occupancy and velocity. How can we design a function that rewards improvements in occupancy and velocity estimates, but can also be computed at test-time using only light curtain placements and measurements? This is challenging because light curtains cannot directly measure velocities; they can only measure the occupancies of a small set of locations where they are placed.


Let us revisit the dynamic Bayes network from Sec.~\ref{sec:background-bf}, shown in Fig.~\ref{fig:dbn}b.
Belief distributions are represented by dynamic occupancy grids. At timestep $t-1$, the grid representing the belief $bel(x_{t-1})$ was forecasted by applying the motion model to obtain the prior belief $\overline{bel}(x_t)$ at timestep $t$ (Eqn.~\ref{eqn:generic-motion-update}, Eqn.~\ref{eqn:dog-motion-update} in App.~\ref{app:dynamic-occupancy-grids}). Then, in the measurement update step, the current light curtain measurement $z_t$ obtained by placing a curtain at locations $u_t$ is used to update the grid to $bel(x_t)$ (Eqn.~\ref{eqn:generic-measurement-update}, Eqn.~\ref{eqn:dog-measurement-update} in App.~\ref{app:dynamic-occupancy-grids}).
Therefore, \textit{we attribute the accuracy of $bel(x_t)$ to action $u_t$}.

The \textit{forecasted occupancy} at time $t+1$ is computed by using the current velocity estimates to forecast the current occupancy by an interval $\Delta t$ using the motion update step
(Eqn.~\ref{eqn:generic-motion-update}, Eqn.~\ref{eqn:dog-motion-update} in App.~\ref{app:dynamic-occupancy-grids}).
The \textit{forecasted occupancy} will be accurate if both the current velocities and current occupancies are accurate. Therefore, the accuracy of forecasted occupancy acts as an appropriate reward function that captures both occupancy and velocity accuracies.

How do we evaluate forecasted occupancy computed using $bel(x_t)$, without requiring ground truth, in a self-supervised way? This is possible by reusing intermediate quantities output during recursive Bayesian updates.

First, note that the forecasted occupancy of $bel(x_t)$ is $\overline{bel}(x_{t+1})$ computed by the next motion update step. {\it Our main insight is that before applying the next measurement update step, $\overline{bel}(x_{t+1})$ can be evaluated using the partial occupancy observed by the next light curtain measurements $z_{t+1}$}. We use the \fone-score between the forecasted occupancy grid and the partially observed occupancy grid \textit{as a self-supervised reward for the previous light curtain placement $u_t$} (See Fig.~\ref{fig:dbn}b). Specifically, we compute the self-supervised reward $R_t = \text{\fone}(\overline{bel}(x_{t+1}), z_{t+1})$, where $\overline{bel}(x_{t+1})$ is computed using Eqn.~\ref{eqn:generic-motion-update} (more specifically, Eqn.~\ref{eqn:dog-motion-update} in App.~\ref{app:dynamic-occupancy-grids}), and $z_{t+1}$ is the partial occupancy observed at time $t+1$. See App.~\ref{app:evaluation-metric} for details on the \fone-score.


An advantage of our self-supervised reward is that it does not require any extra computation. This is because (1) occupancy forecasting of $bel(x_t)$ is performed anyway as part of the motion update step, and (2) the partial occupancy information from $z_{t+1}$ is computed anyway in the next measurement update step. By reusing quantities already computed during recursive Bayes filtering, our self-supervised reward does not require any extra forecasting steps nor  any extra light curtain placements. 

At each timestep, we use the $\epsilon$-greedy strategy to select one among the four curtain placement strategies $a \in \{a_1, a_2, a_3, a_4\}$. Then we compute the curtain placement $u_t$ according to strategy $a$. When the accuracy of the forecasted occupancy $R_t$ is obtained in the next timestep, we update the Q-value of $a$ as $Q(a) \coloneqq Q(a) + \alpha~[R_t - Q(a)]$. We use the \textit{non-stationary reward} formulation~\cite{sutton2018reinforcement} of multi-armed bandits with smoothing parameter $\alpha$ to account for the possibility that different strategies $\{a_1, a_2, a_3, a_4\}$ may be superior at different times. See \app~\ref{app:non-stationary-rewards} for more details.

\section{Parallelized pipeline}
\label{sec:parallelized-pipeline}


Fig.~\ref{fig:parallelized-pipeline} shows our pipeline that has three processes: (1) light curtain sensing, (2) Bayes filtering using dynamic occupancy grids, and (3) computing  curtain placement.
The processes are run in parallel threads with shared memory, at their own independent speeds.

\textbf{1. Light curtain imaging:}
This thread continuously places curtains at locations determined by one of the four strategies  described in Sec.~\ref{sec:intelligent-curtain-strategies}. However, when waiting for the next curtain placement to be computed, it places \textit{random} curtains~\cite{ancha2021rss} (that are generated offline) to sense random locations in the scene. This ensures that the device is always kept busy and runs at approximately 45 Hz.

\textbf{2. Bayes filtering:} This thread inputs light curtain measurements and updates the dynamic occupancy grid. It alternates between motion and measurement update steps (Eqns.~\ref{eqn:generic-motion-update}, \ref{eqn:generic-measurement-update}, Eqns.~\ref{eqn:dog-motion-update}, \ref{eqn:dog-measurement-update} in App.~\ref{app:dynamic-occupancy-grids}). The motion update step requires two grids, each representing the current and next timesteps. Particles are sampled from the \textit{current} grid, perturbed according to the motion model, and inserted into the \textit{next} grid. The roles of the two grids are swapped at every successive motion update to avoid copying data. This thread runs at approximately 35 Hz.

\textbf{3. Computing curtain placement: } This uses the most recent dynamic occupancy grid to compute the next curtain placement (Sec. \ref{sec:intelligent-curtain-strategies}). It first \textit{forecasts} the grid, using the same motion update step (Eqn.~\ref{eqn:generic-motion-update}, Eqn.~\ref{eqn:dog-motion-update} in App.~\ref{app:dynamic-occupancy-grids}), to the next timestep when the next curtain is expected to be imaged. The forecasted occupancy is used to compute the curtain placement. In \app~\ref{app:extra-grid}, we describe how an extra grid is used to ensure thread-safety and that no thread ever needs to wait on another to finish processing. Finally, the control points of the computed curtain are sent to the light curtain device. The three inter-dependent processes are tightly coupled and continuously interact with each other. 

\newcommand{\emphasis}[1]{\color{blue}#1}
\newcolumntype{I}{!{\vrule width 2.5pt}}

\begin{table*}
    \hspace*{-1cm}
    \centering
    \small
    \begin{tabular}{?c?c|c|c|c|cIc|c|c|c|c?}
     \Xhline{1.2pt}
     & \multicolumn{5}{cI}{\textbf{\textit{Simulated environment}}} & \multicolumn{5}{c?}{\textbf{\textit{Real environment}}}\\
     \Xcline{2-11}{1.2pt}
      & \makecell{Classification\\accuracy} & Precision & Recall & \emphasis{\fone-score} & \emphasis{IOU} & \makecell{Classification\\accuracy} & Precision & Recall & \emphasis{\fone-score} & \emphasis{IOU}\\
       \cline{2-11}
      & $\uparrow$ &  $\uparrow$ & $\uparrow$ & \emphasis{$\uparrow$} & \emphasis{$\uparrow$} & $\uparrow$ &  $\uparrow$ & $\uparrow$ & \emphasis{$\uparrow$} & \emphasis{$\uparrow$}\\
      \Xhline{1.2pt}
     \makecell{Simulated\\LiDAR} & 0.9584 & 0.2498 & 0.1073 & \emphasis{0.1360} & \emphasis{0.0787} & \multicolumn{5}{c?}{\it NA}\\
      \hline
      \makecell{Random curtains\\only} & 0.9662 & 0.2698  & 0.0567 & \emphasis{0.0850} & \emphasis{0.0468} & 0.9852 & 0.6306 & 0.2357 & \emphasis{0.2405} & \emphasis{0.2136}\\
       \hline
       \makecell{Max. depth\\probability} & 0.9610 & 0.2448 & 0.1146 & \emphasis{0.1388} & \emphasis{0.0792} & 0.9832 & 0.5943 & 0.2727 & \emphasis{0.3047} & \emphasis{0.2353}\\
      \hline
      \makecell{Max. occupancy\\uncertainty} & 0.9609 & 0.2717 & 0.1266 & \emphasis{0.1493} & \emphasis{0.0857} & 0.9811 & 0.5733 & 0.3041 & \emphasis{0.3319} & \emphasis{0.2515}\\
      \hline
      \makecell{Max. velocity\\uncertainty} & \textbf{0.9648} & 0.2728 & 0.0581 & \emphasis{0.0838} & \emphasis{0.0458} & \textbf{0.9864} & 0.6221 & 0.2615 & \emphasis{0.2545} & \emphasis{0.2232}\\
      \hline
      \makecell{Max. occupancy + velocity\\uncertainty} & 0.9629 & \textbf{0.3026} & 0.1251 & \emphasis{0.1544} & \emphasis{0.0895} & 0.9822 & 0.5899 & 0.3175 & \emphasis{0.3421} & \emphasis{0.2727}\\
      \Xhline{1.2pt}
      \makecell{Multi-armed bandits\\(\textbf{Ours})} & 0.9623 & 0.2814 & \textbf{0.1402} & \emphasis{\textbf{0.1690}} & \emphasis{\textbf{0.0976}} & 0.9854 & \textbf{0.6467} & \textbf{0.3647} & \emphasis{\textbf{0.3703}} & \emphasis{\textbf{0.3053}}\\
      \Xhline{1.2pt}
    \end{tabular}
    \caption{Accuracy of occupancy and velocity estimation measured using \textit{forecasted occupancy} in (a) simulated, and (b) real environments.}
    \label{table:ve-results}
\end{table*}

\section{Experiments}
\label{sec:experiments}

\begin{figure}
    \centering
    \adjustbox {padding=-4pt 0pt -4pt 0pt,margin=0pt 0pt 0pt 0pt,frame=0pt 0pt}
    {
        \includegraphics[trim=0 0 0 0,clip,width=0.5\textwidth]{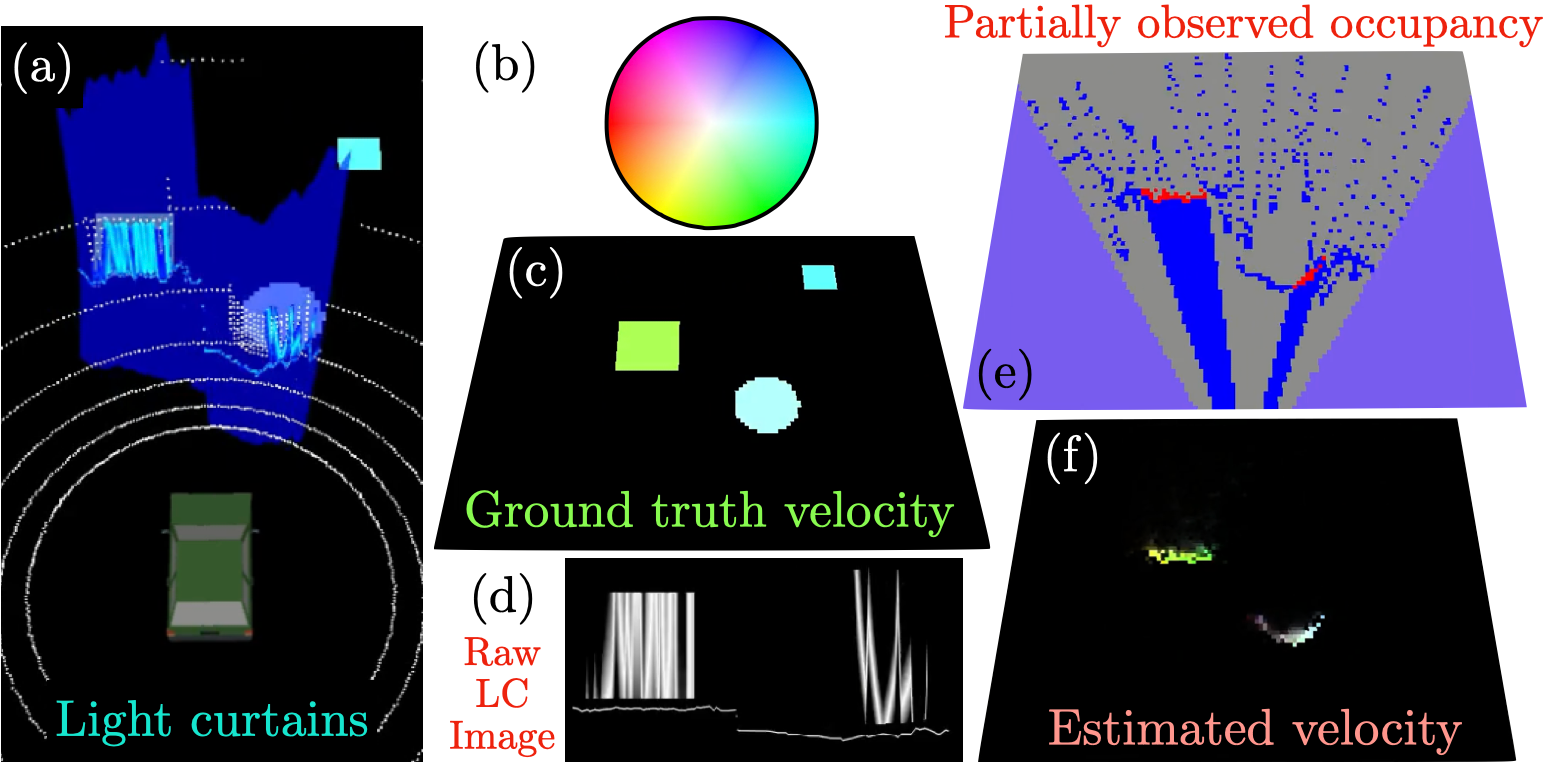}
    }
    \caption{\textit{Velocity estimation in simulation.}
    (a) The environment with moving blocks and curtain placement shown in blue.
    (b) Color-coding for visualizing velocity from the top-down view.
    (c) Ground truth occupancy and velocity.
    (d) Raw light curtain images; high intensities are good because it means that object surfaces were found and intersected by the light curtain.
    (e) Partial occupancy observations from light curtain measurements that are input to the dynamic occupancy grid.
    (f) Velocity and occupancy estimated by the dynamic occupancy grid.
    We advise the reader to view video examples on the \href{\website}{project website}.}
    \label{fig:qualitative-sim}
\end{figure}

\begin{figure*}[h!]
    \centering
    \includegraphics[trim=0 0 0 0,clip,width=1.0\textwidth]{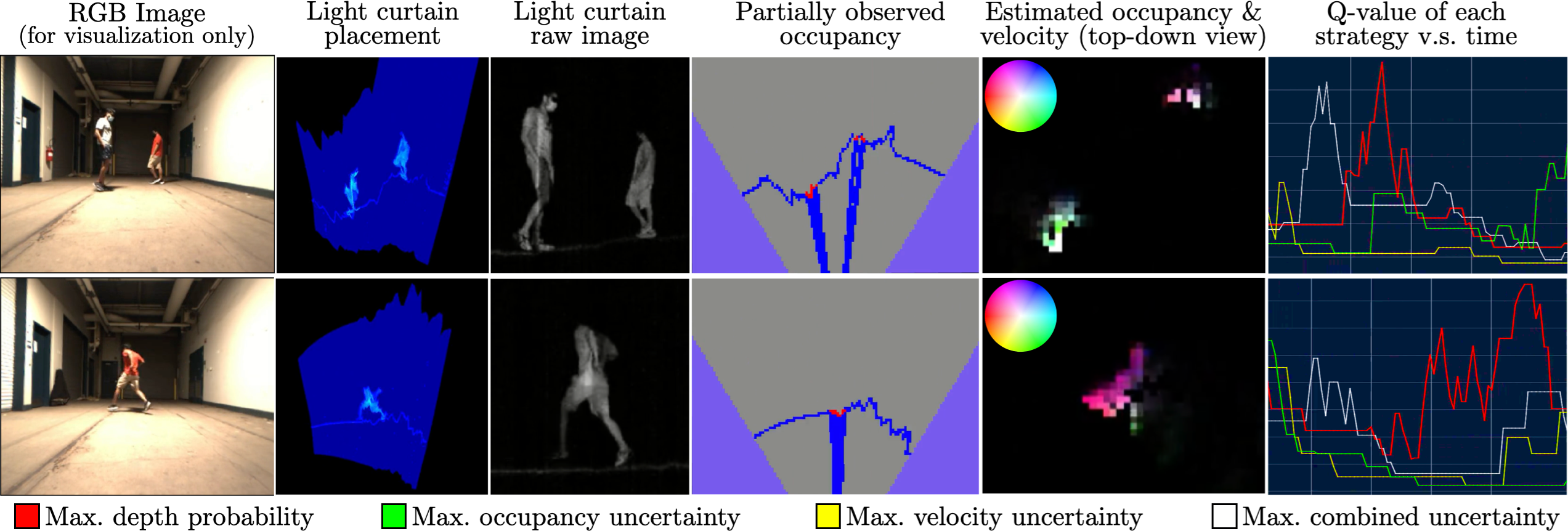}
    \caption{\textit{Velocity estimation in the real world using multi-armed bandits (MAB)}. Please refer to titles and Fig.~\ref{fig:qualitative-sim} for descriptions of each column.
    Our method only uses light curtains; RGB images are for visualization only. The rightmost column shows the Q-values of each strategy. Higher Q-value is better; the action with the highest Q-value is chosen during exploitation.
    \textit{Top row:} shows two pedestrians walking at relaxed speeds. The directions of motion are correctly inferred for each person: the pedestrian walking to the right is shown in greenish-blue and the person walking to the left is colored in red. The current action selected maximizes occupancy uncertainty. The bottom row shows a more challenging environment where where a lone pedestrian performs fast motion: running and jumping. The direction of velocity is correctly inferred as moving top-left (left and away from the sensor) i.e. reddish pink. The color saturation is high indicating the larger magnitude of velocity. The current action selected maximizes depth probability. We advise the reader to view the video examples on the \href{\website}{project website}.}
    \label{fig:qualitative-real}
\end{figure*}

\subsection{Environments}

\noindent \textbf{Simulation environment}: We use a simulated environment consisting of various blocks moving in a variety of motions (see Fig.~\ref{fig:wheelchair-envs}b). The environment contains cylinders and cuboids, moving in (1) linear, harmonic (oscillatory) motion along different directions, (2) curved sinusoidal motion, and (3) random Brownian motion.
We use an efficient light curtain simulator described in App.~\ref{app:efficient-simulation}.

\noindent \textbf{Real-world environment}: Our real-world environment consists of a mobile robot with a mounted light curtain device (Fig.~\ref{fig:wheelchair-envs}a) navigating in the presence of two pedestrians walking in multiple directions, at different speeds and in complicated trajectories (see Fig.~\ref{fig:wheelchair-envs}c).

\subsection{Evaluation metrics}
\label{sec:evaluation-metric}

Since we wish to evaluate the accuracy of both occupancy and velocity estimates, we use the \textit{forecasted occupancy}~\cite{mahjourian2022occupancy,waymo2022challenge} as our evaluation metric. As noted in Sec.~\ref{sec:ss-rewards}, the forecasted occupancy will be accurate if both current velocities and current occupancies are accurate.
The future occupancy at time $t + \Delta t$ is computed by the motion update step (Eqn.~\ref{eqn:generic-motion-update}, Eqn.~\ref{eqn:dog-motion-update} in App.~\ref{app:dynamic-occupancy-grids}) that uses the current velocity to forecast the current occupancy by an interval $\Delta t$.
This metric is particularly relevant for obstacle avoidance where estimates of future obstacle locations must be accurately computed to plan safe, collision-free paths.

Ideally, the accuracy of forecasted occupancy can be computed by comparing it against ground truth occupancy at $t + \Delta t$.
This is possible in simulated environments where ground truth occupancy is available for all grid cells. In real-world environments, true occupancy can only be measured for a subset of cells by the light curtain; in this case, we use the ``self-supervised'' version of the metric described in Sec.~\ref{sec:ss-rewards}.
We follow prior works~\cite{hornung2013octomap,meyer2012occupancy,tatarchenko2017octree} that treat the evaluation of occupancy as a classification problem and compute several metrics: (1) classification accuracy~\cite{hornung2013octomap,meyer2012occupancy}, (2) precision, (3) recall, (4) \fone-score and (5) the IoU~\cite{tatarchenko2017octree} between the predicted and ground truth occupancy masks. For more details on these metrics, please see \app~\ref{app:evaluation-metric}.

\subsection{Quantitative analysis}

Table~\ref{table:ve-results} shows the performance of various light curtain placement strategies in simulated and real-world environments, evaluated using multiple forecasted occupancy metrics (see Sec.~\ref{sec:evaluation-metric}, \app~\ref{app:evaluation-metric}).
Since a large proportion of cells are unoccupied, the classification accuracy of all methods is very similar. Furthermore, precision and recall metrics can be deceived by mostly predicting negative and positive labels respectively. However, the \fone-score and IoU metrics are discriminative and robust; they are high only when both precision and recall are high. Therefore, we focus on these two metrics (shown in blue).
In both sets of experiments, multi-armed bandits that combine the four curtain placement policies using our self-supervised reward outperform all other methods. This shows that intelligently switching between multiple placement strategies is more beneficial than using any one single strategy at all times.

Between the other four strategies, maximizing occupancy uncertainty and maximizing a linear combination of occupancy and velocity uncertainty perform comparably.
Maximizing velocity uncertainty tends to perform the worst.
Fortunately, multi-armed bandits learn to downweight this under-performing strategy (see Table~\ref{table:ve-mab-analysis}, rightmost column in Fig.~\ref{fig:qualitative-real}). We also compare against other baselines: using only random curtains (without placing any computed curtains), and with a simulated LiDAR. Unsurprisingly, using random curtains performs the worst. All non-random curtain policies except maximizing velocity uncertainty are able to outperform LiDAR.
This is because light curtains are faster ($\sim$45 Hz) and can be placed intelligently to maximize the accuracy of occupancy and velocity estimates. 

Table~\ref{table:ve-mab-analysis} shows an analysis specific to the multi-armed bandit method. Please see the caption for details.
We find that the best performing policies in Table~\ref{table:ve-results} have the highest Q-values and are selected most frequently. The following trend holds: the better the performance of an individual policy when used in isolation (shown in Table~\ref{table:ve-results}), the higher is its average Q-value and its frequency of being chosen. However, a combination of all policies (MAB) is better than any single one.

\begin{table}
    \centering
    \small
    \begin{tabular}{?c?c|c?}
     \Xhline{1.2pt}
      & \makecell{Frequency\\of selection} & \makecell{Avg. Q-value\\function (IOU)}\\
       \cline{2-3}
      \Xhline{1.2pt}
      \makecell{Max. depth\\probability} & 22.9\% & 0.261\\
       \hline
       \makecell{Max. occupancy\\uncertainty} & 31.1\% & 0.276\\
      \hline
      \makecell{Max. velocity\\uncertainty} & 13.4\% & 0.202\\
      \hline
      \makecell{Max. occupancy + velocity\\uncertainty} & 32.5\% & 0.292\\
      \Xhline{1.2pt}
    \end{tabular}
    \caption{Quantitative analysis of the multi-armed bandit method. The first column shows the percentage of times each action (i.e. curtain placement policy) was chosen. The second column shows the average Q-value of each action computed by the multi-armed bandit. Higher Q-value is better; the action with the highest value is selected during exploitation.}
    \label{table:ve-mab-analysis}
\end{table}

\subsection{Qualitative analysis}
\label{sec:qualitative-analysis}

\noindent \textbf{Visualizing velocities and occupancies:} We use the HSV colorwheel~\cite{HSV} shown in Fig.~\ref{fig:qualitative-sim} and \ref{fig:qualitative-real}, to jointly visualize velocities and occupancies. The color `value' (from HSV) encodes the occupancy probability; dark is low occupancy probability and bright is high occupancy probability. The `hue' encodes the direction of velocity from the top-down view. `Saturation' encodes the magnitude of velocity: white is stationary whereas colorful corresponds to high speed. See \app~\ref{app:colorwheel} for more details.

\textbf{Examples}. Fig.~\ref{fig:qualitative-sim} shows an example of velocity estimation in the simulated environment and Fig.~\ref{fig:qualitative-real} shows qualitative results on the real-world environment using our multi-armed bandits (MAB) curtain placement method. Please see captions for explanation. We advise the reader to view the video examples on the \href{\website}{project website}.
In Fig.~\ref{fig:qualitative-sim}, we see that the estimated velocities appear to be consistent with the ground truth, as shown by the corresponding colors that indicate the estimated and ground-truth velocity directions.

\textbf{Full-stack navigation}. We integrate our system into a full-stack navigation pipeline~\cite{cao2022autonomous} that performs planning, control and obstacle avoidance. We mount the light curtain device on a mobile robot (see Fig.~\ref{fig:wheelchair-envs}a). We use ORB-SLAM3~\cite{campos2021orb} for localization and mapping that takes depth from light curtains as input. Using position and velocity estimates, the robot is able to perform dense mapping in an indoor environment and avoids static and dynamic obstacles. Please see \app~\ref{app:full-stack-navigation} for more details.
\section{Conclusion} 
\label{sec:conclusion}

In this work, we develop a method using programmable light curtains, an actively controllable resource-efficient sensor, to estimate the positions and velocities of objects in complex, dynamic scenes.
We use a probabilistic framework based on particle filters and occupancy grids to estimate velocities from partial light curtain measurements.
We design curtain placement policies that verify predicted object locations and maximize information gain.
Importantly, we combine the strengths of these policies using a novel multi-armed bandits framework that switches between the placement strategies to improve performance. This is enabled by our novel self-supervised reward function that evaluates current velocity estimates using future light curtain placements with only minimal computational overhead. We integrate our method into a full-stack navigation system that performs localization, mapping and obstacle avoidance using light curtains. We hope our work paves the way for combining multiple sensor control strategies using self-supervised feedback for  perception and navigation in complex and dynamic environments.

\section*{Acknowledgments}
We thank Pulkit Grover for discussions on information-theoretic measures of mixed discrete-continuous random variables. This material is based upon work supported by the National Science Foundation under Grants No. IIS-1849154, IIS-1900821, the United States Air Force and DARPA under Contract No. FA8750-18-C-0092, and a grant from the Manufacturing Futures Institute at Carnegie Mellon University.


\bibliographystyle{plainnat}
\bibliography{main}

\clearpage

\begin{appendices}
  
  \begingroup
  \centering
  \twocolumn[
    \centering
    {\fontsize{28pt}{28pt}\selectfont Appendix}\\
    \vspace{1em}
    {\fontsize{20pt}{20pt}\selectfont Active Velocity Estimation using Light Curtains\\via Self-Supervised Multi-Armed Bandits}
    \vspace{2em}
  ]
  \endgroup

  \section{Derivation of recursive Bayesian estimation}
\label{app:bayes-filtering}

\noindent The goal is to infer at each timestep $t$ a distribution $bel(x_t) = P(x_t \mid u_{1:t}, z_{1:t})$ over the current state $x_t$ from the sequence of sensor observations $z_{1:t}$ and the known sequence of actions $u_{1:t}$.
$bel(x_t)$ is computed using \textit{recursive Bayesian estimation}~\cite{thrun2002probabilistic}. Combining the definition of $bel(x_t)$ and the Markov property of the dynamic Bayes network, we can derive the following recursive relationship~\cite{thrun2002probabilistic}:
\begin{align*}
    &bel(x_t) = P(x_t \mid u_{1:t}, z_{1:t})\\
        &\hspace{2em} \propto P(x_t, z_t \mid u_{1:t}, z_{1:{t-1}})\\
        &\hspace{2em} = P(z_t \mid x_t, u_{1:t}, z_{1:{t-1}})~P(x_t \mid u_{1:t}, z_{1:{t-1}})\\
        &\hspace{2em} = P(z_t \mid x_t, u_t)~P(x_t \mid u_{1:{t-1}}, z_{1:{t-1}})\\
        &\hspace{2em} = P(z_t \mid x_t, u_t)~\int_{x_{t-1}} \hspace{-1em} P(x_{t-1}, x_t \mid u_{1:{t-1}}, z_{1:{t-1}})~\del x_{t-1}\\
        &\hspace{2em} = P(z_t \mid x_t, u_t)~\int_{x_{t-1}} \hspace{-1em} P(x_{t-1} \mid u_{1:{t-1}}, z_{1:{t-1}})\\
        &\hspace{2em} \phantom{= P(z_t \mid x_t, u_t)~\int_{x_{t-1}} \hspace{-1em} } \cdot P(x_t \mid x_{t-1}, u_{1:{t-1}}, z_{1:{t-1}})~\del x_{t-1}\\
        &\hspace{2em} = P(z_t \mid x_t, u_t)~\int_{x_{t-1}}  \underbrace{P(x_{t-1} \mid u_{1:{t-1}}, z_{1:{t-1}})}_{bel(x_{t-1})}\\
        &\hspace{2em} \phantom{= P(z_t \mid x_t, u_t)~\int_{x_{t-1}}} \cdot P(x_t \mid x_{t-1})~\del x_{t-1}\\
        &\hspace{2em} = P(z_t \mid x_t, u_t)~\int_{x_{t-1}} \hspace{-1em} bel(x_{t-1})~P(x_t \mid x_{t-1})~\del x_{t-1}\\
        &\hspace{2em} = P(z_t \mid x_t, u_t)~\overline{bel}(x_t) \text{\ \color{blue}(Measurement update)} \text{,\ where}\\
    &\overline{bel}(x_t) = \int_{x_{t-1}} \hspace{-1em} bel(x_{t-1})~P(x_t \mid x_{t-1})~\del x_{t-1} \text{\ \color{blue}(Motion update)}
\end{align*}

Based on the above recursive equations, recursive Bayesian estimation alternates between the following two steps:
\begin{enumerate}
    \item \textbf{Motion update step}: This step accounts for the dynamics of the environment. It first computes an intermediate quantity defined above:\\$\overline{bel}(x_t) = \int_{x_{t-1}} bel(x_{t-1})~P(x_t \mid x_{t-1})~\del x_{t-1}$. This is the result of ``applying'' a known or assumed motion model $P(x_t \mid x_{t-1})$ to the previous belief $bel(x_{t-1})$. In dynamic occupancy grids, this step accounts for the motion of scene points based on their current 2D velocities. The occupancies and velocities of the next timestep are computed based on the occupancies and velocities in the previous timestep. We use a constant velocity motion model with Gaussian noise in both velocity and position. When the motion model is applied to Fig.~\ref{fig:dynamic-occupancy-grid}a, it is updated to Fig.~\ref{fig:dynamic-occupancy-grid}b (illustration only). This correction step usually increases the uncertainty in occupancies and velocities.
    \item \textbf{Measurement update step:} This step incorporates measurements from a sensor. It updates the prior belief $\overline{bel}(x_t)$ to $bel(x_t) \propto P(z_t \mid x_t, u_t)~\overline{bel}(x_t)$ by weighting $\overline{bel}(x_t)$ by the likelihood of the observed measurements $P(z_t \mid x_t, u_t)$. This step usually reduces uncertainty in the state. In dynamic occupancy grids, occupancies are updated using the measurements from the light curtain. Since light curtains (or any depth sensor) only measure the locations of objects, this step does not update velocities. The velocity estimates are automatically refined in subsequent motion update steps. The measurement update reduced the probability of one of the positions of an object in Fig.~\ref{fig:dynamic-occupancy-grid}b to Fig.~\ref{fig:dynamic-occupancy-grid}c (illustration only). 
\end{enumerate}
  \section{Dynamic occupancy grids}
\label{app:dynamic-occupancy-grids}

\begin{figure*}[th!]
    \centering
    \subfloat[\small Dynamic occupancy grid before the motion update step.]
    {
        \hspace{0.5em}
        \includegraphics[trim=0 0 0 0,clip,width=0.47\textwidth]{figures/dog/grid.png}
        \hspace{0.5em}
    }\\
    \subfloat[\small After the motion update step.]
    {
        \hspace{0.5em}
        \includegraphics[trim=-60 0 -80 0,clip,width=0.3\textwidth]{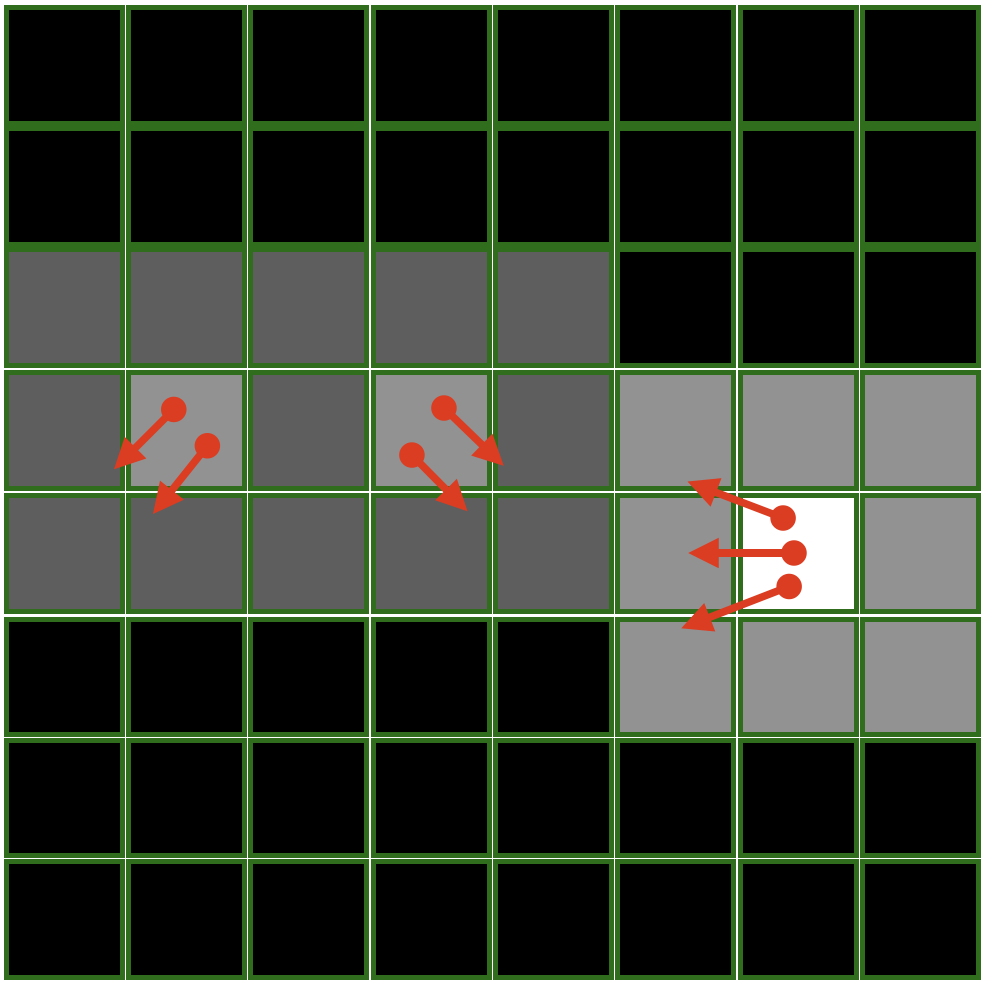}
        \hspace{0.5em}
    }
    \subfloat[\small After the measurement update step.]
    {
        \hspace{0.5em}
        \includegraphics[trim=-80 0 -60 0,clip,width=0.3\textwidth]{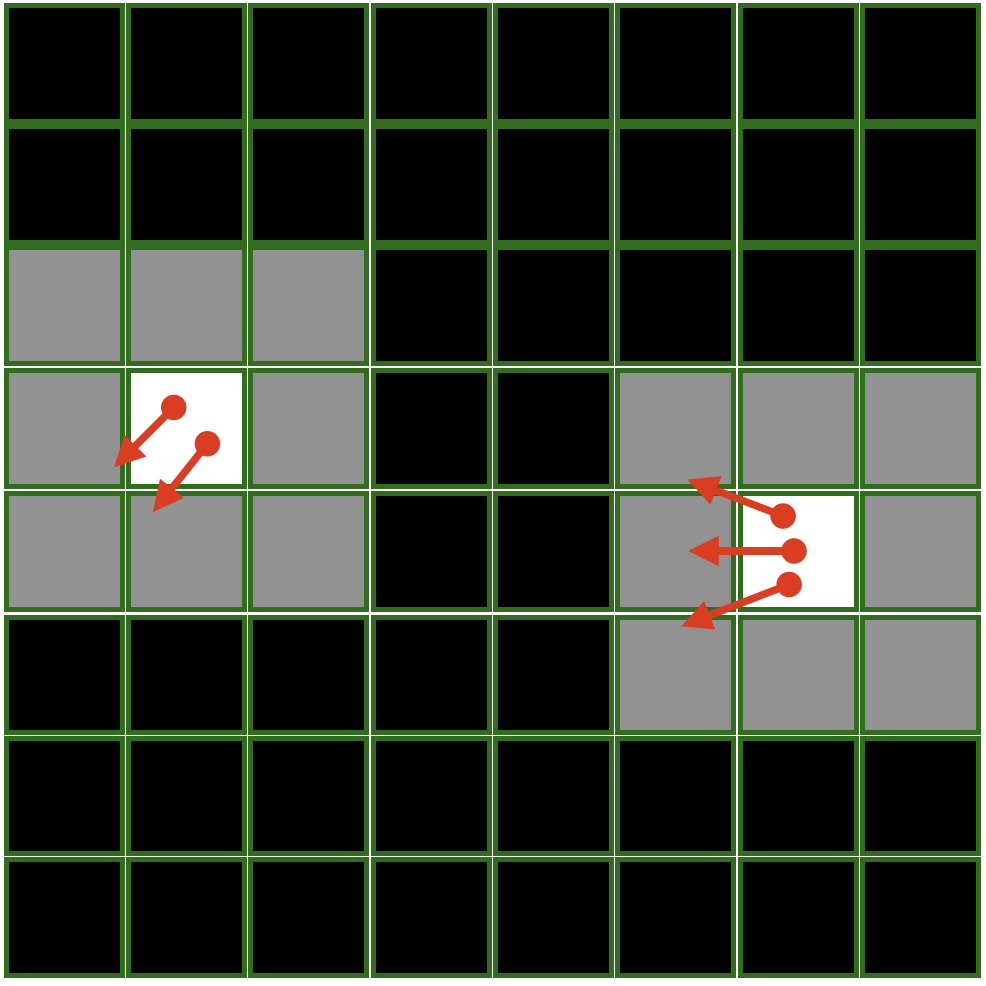}
        \hspace{0.5em}
    }
    \caption{\textit{Dynamic occupancy grid.} \textbf{Grid structure:} The 2D grid represents the top-down view and is made up of cells. Like conventional static occupancy grids~\cite{elfes1989using,thrun2002probabilistic}, each cell contains an occupancy probability $p \in [0, 1]$. Dark indicates low occupancy probability and bright indicates high occupancy probability. In addition, each cell also contains a set of weighted \textit{particles} where each particles stores a single 2D velocity. The set of particles together represents a probability distribution of that cell's velocity. \textbf{Grid updates:} The grid is a Bayes filter~\cite{thrun2002probabilistic} that consists of two steps: the motion update step and the measurement update step. (a) The grid before performing any update. (b) \textit{Motion update step}: the occupancies and velocities of the next timestep are computed based on the grid in the previous timestep, increasing uncertainty. (c) \textit{Measurement update step:} the occupancies are updated using the measurements from the light curtain, decreasing uncertainty. This will refine the velocity estimates.}
    \label{fig:dynamic-occupancy-grid}
\end{figure*}

\noindent The dynamic occupancy grid, like conventional occupancy grids~\cite{elfes1989using,thrun2002probabilistic}, is an instance of Bayes filters.
Occupancy grids~\cite{elfes1989using,thrun2002probabilistic} are a standard tool in robotics for mapping the location of static objects in the environment. 2D occupancy grids that map objects from the top-down view are commonly used for mapping and SLAM in robot navigation. Each cell in the grid contains an \textit{occupancy probability} $p\in [0, 1]$, denoting the probability of the cell being occupied by an object.
\textit{Dynamic} occupancy grids~\cite{danescu2011modeling} are an extension of classical occupancy grids (see Figure~\ref{fig:dynamic-occupancy-grid}a). Each cell in the grid contains both (1) the occupany probability $p \in [0, 1]$, as well as (2) a probability distribution over 2D velocities. The velocity distribution is represented by a set of weighted particles, where each particles stores a single 2D velocity. The set of weighted particles approximates the true velocity distribution.

\subsection{Mathematical framework}
\label{app:mathematical-framework}

Our method is built upon dynamic occupancy grids introduced by Danescu. et. al.~\cite{danescu2011modeling}. The authors describe particles as both representing a velocity distribution (i.e. weighted velocity hypotheses), as well as being ``physical building blocks of the world". The former interpretation suggests that particles together represent the probability distribution of the velocity of a single physical scene point, whereas the latter suggests that each particle corresponds to its own scene point. Furthermore, the particles not only represent velocities, but their count represents the probability of occupancy. While the method is shown to be very promising, the precise role of particles and what they represent remains unclear. In this work, we re-derive dynamic occupancy grids using a more rigorous mathematical analysis. We explicitly state the assumptions made and provide a precise interpretation of particles. Our framework can be derived from three reasonably mild assumptions:

\begin{tcolorbox}[colback=gray!30, colframe=black!60, title=\textbf{Assumption 1: There are no collisions}]
    \textit{Each cell can be occupied by at most one physical scene point with a single velocity. Cells are sufficiently small that multiple objects with different velocities cannot exist (``collide'') within a cell.}
\end{tcolorbox}

This assumption is required for a single velocity of a cell to be well-defined. Assumption 1 paves the way for a straightforward interpretation particles: all particles belonging to a cell represent a probability distribution over the single velocity of that cell. Assumption 1 allows us to define the state space of occupancies and velocities.

\textbf{Representing the state space}: Each cell is indexed by $i \in \mathcal{I}$ from an index set $\mathcal{I}$ of all cell in the grid. At timestep $t$, the state of the $i$-th cell is denoted by $x^i_t = (o^i_t, v^i_t)$. It contains two variables. The first is a binary occupancy variable $o^i_t \in \{0, 1\}$ which denotes whether the cell is occupied or not. The second is a 2D velocity variable $v^i_t \in \mathbb{R}^2$ representing the continuous velocity of the cell (\textit{if} it is occupied) from the top-down view. The overall state of the dynamic occupancy grid $x_t$ is a concantenation of the states of all cells in the grid, i.e. $x_t = \{x^i_t = (o^i_t, v^i_t) \mid i \in \mathcal{I}\}$. Note that the variable $v^i_t$ is ``conditional'': it is only defined when the cell is occpied i.e. $o^i_t = 1$.

\begin{tcolorbox}[colback=gray!30, colframe=black!60, title=\textbf{Assumption 2: Constant velocity motion model}]
    \textit{Each scene point moves with a constant velocity, with added Gaussian noise.}
\end{tcolorbox}

Any motion can be approximated by a constant velocity motion model as long as the time interval is sufficiently small. Therefore, this assumption is reasonable in our setting since light curtains operate at very high speeds (45-60 Hz). Note that although we use the constant velocity motion model in this work following \cite{danescu2011modeling}, the dynamic occupancy grid framework can still be used by swapping it with any other motion model of choice.

Let $\epsilon^i_t \sim \mathcal{N}(0, R_\epsilon)$ and $\delta^i_t \sim \mathcal{N}(0, R_\delta)$ be the Gaussian noise in velocity and position respectively for the $i$-th cell in the grid at time $t$. And let $\pos^i$ denote the 2D location of the center of the $i$-th cell. The constant velocity motion model can be expressed mathematically as

\begin{align}
    o^j_{t+1}, v^j_{t+1} =
    \begin{cases}
        1,\ v^i_t + \epsilon^i_t\ \ \ \mathrm{if\ } \exists i \in \mathcal{I} \text{ such that } o^i_t = 1 \text{ and}\\
        \phantom{1,\ v^i_t + \epsilon^i_t\ \ \ \mathrm{if\ } \exists i} \pos^j \approx \pos^i + v^i_t~\Delta t + \delta^i_t\\
        0 \phantom{,\ v^i_t + \epsilon^i_t\ \ \ } \text{otherwise}
    \end{cases}
    \label{eqn:state-space-motion-model} 
\end{align}

In other words, a cell $j$ will be occupied in the next timestep if and only if there exists another cell $i$ in the previous timestep which moves to $j$ under the constant velocity motion model. In that case, the velocity of cell $j$ will be equal to that of cell $i$ modulo the Gaussian noise. The equality is approximate taking into account the finite size of the cell.

\textbf{Representing the belief distribution}: A dynamic occupancy grid represent the current belief over velocities and occupancies of the environment. It is a probability distribution over states $x_t$ described above. The state space is extremely large. Dynamic occupancy grids represent the belief compactly by making the following assumption:
{
\hspace{-10em}
\begin{tcolorbox}[colback=gray!30, colframe=black!60,width=0.47\textwidth,
title=\textbf{Assumption 3: Cells are mutually independent},]
    \textit{The probability distributions of all cells are mutually independent. This is a standard assumption made for occupancy grids for computational tractability.}
\end{tcolorbox}
}
Each cell contains two distributions:
\begin{itemize}
    \item \underline{Occupancy distribution} ($\omega^i_t$): $o^i_t$ is a Bernoulli random variable over $\{0, 1\}$ with probability $\omega^i_t \in [0, 1]$.
    \item \underline{Velocity distribution} ($V^i_t$): $v^i_t$ is a random variable over $\mathbb{R}^2$. It is represented by a set of $M$ weighted particles $V^i_t = \{(v^{i, m}_t, p^{i, m}_t) \mid 1 \leq m \leq M \}$. The $m$-th particle has a velocity $v^{i, m}_t$ and weight $p^{i, m}_t$ that add up to 1 i.e. $\sum_{m=1}^M p^{i, m}_t = 1$. The larger the number of particles used, the better is the particle approximation to the true continuous velocity distribution.
\end{itemize}
The velocity distribution acts like a ``conditional'' distribution. The probability that cell $i$ is occupied with velocity $v^{i, m}_t$ is the product $\omega^i_t~p^{i, m}_t$ of the probability of being occupied ($\omega^i_t$) and the probability of having the velocity $v^{i, m}_t$ given that it is occupied ($p^{i, m}_t$). The probability of being unoccupied is simply $1-\omega^i_t$. Since cells are assumed to be mutually independent, the probability of the entire grid is the product of probabilities of individual cells in the grid.

\subsection{Motion update step}
\label{app:motion-update-step}

\begin{figure}[b!]
    \centering
    \includegraphics[trim=0 0 0 0,clip,width=0.3\textwidth]{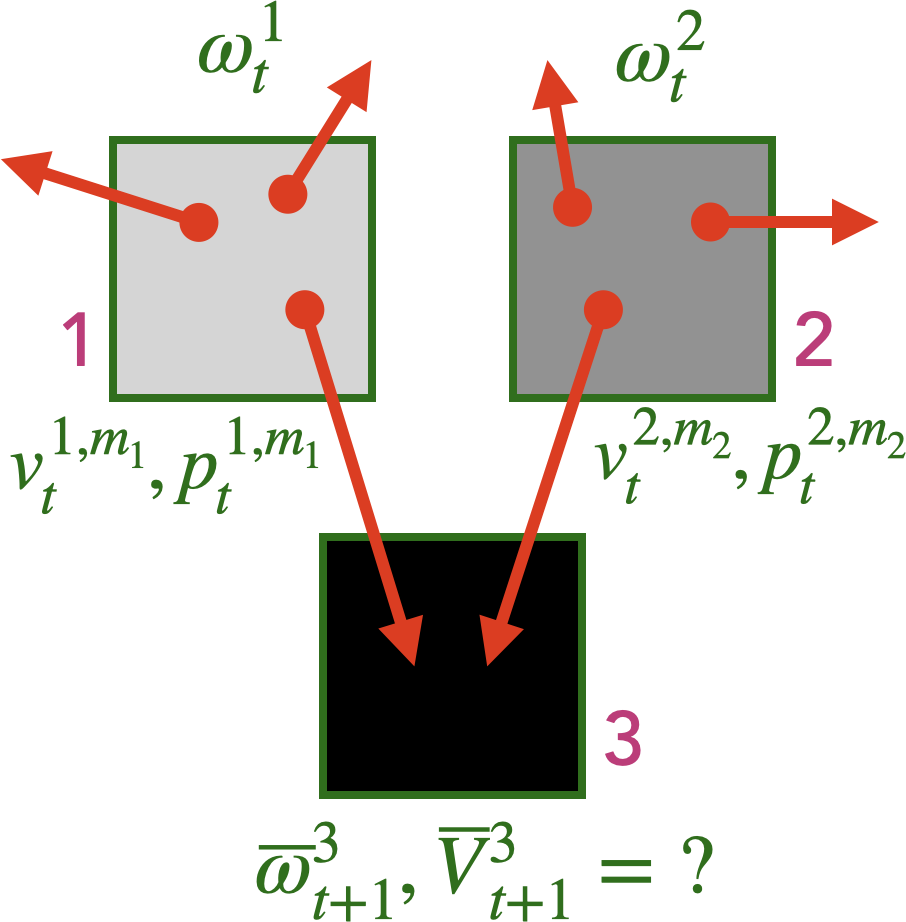}
    \caption{Applying the motion model to update occupancies and velocities of cells in the motion update step. $\omega$'s are occupancy probabilities of cells, $v$'s and $p$'s are velocities and weights of the individual particles respectively.}
    \label{fig:motion-update}
\end{figure}

We will now derive the motion update equations of the dynamic probability grid. These equations govern how particles will move across the grid and be reweighted. Consider a simplified grid shown in Fig.~\ref{fig:motion-update}a. Assume that there are only three cells in the grid, and only cells 1 and 2 are occupied at time $t$. The occupancy and velocity distributions are illustrated in the figure, and particles $(v^{1, m_1}_t, p^{1, m_1}_t)$ from cell 1 and $(v^{2, m_2}_t, p^{2, m_2}_t)$ from cells 2 move to cell 3 in the next timestep $t+1$ (assuming that noise has been incorporated in $v^{1, m_1}_t, v^{2, m_2}_t$). What should be the occupancy and velocity distribution of cell 3?

Let $E_1$ be the event that the particle from cell 1 enters cell 3. Similarly, let $E_2$ be the event that the particle from cell 2 enters cell 3. We have that $P(E_1) = \omega^1_t~p^{1, m_1}_t$ and $P(E_2) = \omega^2_t~p^{2, m_2}_t$. Cell 3 will be occupied when either $E_1$ or $E_2$ happens (from Eqn.~\ref{eqn:state-space-motion-model}). Its occupancy probability after the motion update is $\overline{\omega}^3_{t+1} = P(E_1 \cup E_2)$. Now, from assumption 1 (the no collision assumption), objects of different velocities cannot occupy the same cell. Therefore events $E_1$ and $E_2$ must be disjoint. From the law of total probability, if $E_1$ and $E_2$ are disjoint, then $P(E_1 \cup E_2) = P(E_1) + P(E_2)$. Therefore, the occupancy probability of cell 3 after the motion update $\overline{\omega}^3_{t+1} = \omega^1_t~p^{1, m_1}_t + \omega^2_t~p^{2, m_2}_t$. The conditional velocity distribution of cell 3 will comprise of $v^{1, m_1}_t$ and $v^{2, m_2}_t$, with weights proportional to $P(E_1)$ and $P(E_2)$ respectively. This leads us to the general motion update of dynamic occupancy grids:
\begin{equation}
    \textit{\underline{Motion update step for dynamic occupancy grids}}
    \label{eqn:dog-motion-update}
\end{equation}\vspace{-2em}
\begin{align*}
    \hspace{-1em}\overline{\omega}^j_{t+1} &= \sum_{i \in \mathcal{I}} \omega^i_t \sum_{m=1}^{M_i} p^{i, m}_t~\mathbb{I}\big[\pos^j \approx \pos^i + v^{i, m}_t~\Delta t + \delta^{i, m}_t\big]\\
    \hspace{-1em}\overline{V}^j_{t+1} &= \Big\{ \big(v^{i, m}_t + \epsilon^{i, m}_t, \frac{\omega^i_t~p^{i, m}_t}{\overline{\omega}^j_{t+1}} \big)~\Big\vert~ i, m: \pos^j \approx \pos^i + v^{i, m}_t~\Delta t + \delta^{i, m}_t \Big\}
\end{align*}
Note that the above derivation does not require assumption 3; the law of total probability applies even when the distributions of incoming cells are not independent! Therefore, the motion update is exact even if we treat $\omega^i_t$ and $V^i_t$ as marginal distributions of potentially correlated cells. Assumption 3 is however required for the measurement update step (see \app~\ref{app:measurement-update-step}).

When adding the $\omega^i_t~p^{i, m}_t$ terms from incoming particles to compute $\overline{\omega}^j_{t+1}$, the sum should not exceed 1 under the no collision assumption (assumption 1). However, in practice, this may be violated. In such cases, we truncate the occupancy probability to 1 following~\cite{danescu2011modeling}. However, this happens rarely; we needed to perform truncation only 0.35\% times on average.

\subsection{Measurement update step}
\label{app:measurement-update-step}

Sensors such as light curtains and LiDARs provide depth information, from which the current occupancy of cells can be inferred. We use a post-processing algorithm described in \cite{hu2020you} to process sensor data and output a detection variable $z^i_t \in \{ \texttt{OCCUPIED}, \texttt{FREE}, \texttt{UNKNOWN} \}$ for each cell $i$ in the grid. $z^i_t$ indicates the presence, absence or lack of knowledge about objects inside cell $i$.

For LiDAR scans, \citet{hu2020you} marks each cell that contains LiDAR points as \texttt{OCCUPIED}. Then, it uses the fact that if a 3D point was detected, light must have traveled between the sensor and the detected point in a straight line without obstruction. Therefore, rays are cast starting from the sensor to the \texttt{OCCUPIED} cells using an efficient voxel traversal algorithm~\cite{amanatides1987fast}. Cells lying along these rays are marked \texttt{FREE}. Any cells that remain unclassified are marked as \texttt{UNKNOWN}. This method exploits visibility constraints of light to extract the maximum possible information from a 3D scan. We use the same processing method customized for light curtains. When a light curtain is placed on a set of cells in the grid, the cells are classified as \texttt{OCCUPIED} or \texttt{FREE} based on whether points were detected inside the cell. Raycasting to occupied cells is also performed to discover additional \texttt{FREE} cells. Figure~\ref{fig:dog}b visualizes an example of visibility classification. Cells detected as \texttt{OCCUPIED} are shown in red.  Cells shown in blue are inferred as \texttt{FREE} because they either lie undetected on the curtain or lie on rays cast to red cells. \texttt{UNKNOWN} cells are shown in gray.

The measurement update step takes the visibility classification as input. Our observation model treats this classification as a noisy observation of true occupancy. We do not update the occupancy of \texttt{UNKNOWN} cells. For known cells, we assume a false positive rate $\fprate \in [0, 1]$ and a false negative rate $\fnrate \in [0, 1]$. The observation model is:
\begin{align*}
    P(z^i_t = \texttt{OCCUPIED} \mid o^i_t = 1) &= 1 - \fnrate\\
    P(z^i_t = \texttt{FREE} \mid o^i_t = 1) &= \fnrate\\
    P(z^i_t = \texttt{OCCUPIED} \mid o^i_t = 0) &= \fprate\\
    P(z^i_t = \texttt{FREE} \mid o^i_t = 0) &= 1-\fprate
\end{align*}
We first use the assumption that all cells are mutually independent (assumption 3) to write the belief of the overall grid $\overline{bel}(x_t) = \prod_{i \in \mathcal{I}} \overline{bel}(x^i_t)$ as a product of belief distributions of each cell in the grid. Since the likelihood function $P(z_t \mid x_t) = \prod_{i \in \mathcal{I}} P(z^i_t \mid o^i_t)$ is also independent for each cell, the updated posterior belief $bel(x_t \mid z_t) \propto \overline{bel}(x_t)~P(z_t \mid x_t)$ can be computed independently for each cell. Given the prior occupancy distribution $\overline{\omega^i_t}$ and an observation $z^i_t$ for cell $i$, its occupancy distribution after the measurement update can be computed using the Bayes rule:
\begin{align}
    \textit{Measurement update step for dynamic occupancy grids}
    \label{eqn:dog-measurement-update}\\
\omega^i_t = \frac{\overline{\omega^i_t}~P(z^i_t \mid o^i_t=1)}{\overline{\omega^i_t}~P(z^i_t \mid o^i_t=1) + (1-\overline{\omega^i_t})~P(z^i_t \mid o^i_t=0)} \nonumber
\end{align}
Depth sensors only provide information about the occupancy of cells; they do not directly measure object velocities. Velocities are inferred indirectly by a combination of measurement- and motion- update steps. The measurement update incorporates information about occupancy and the motion update infers velocities that are consistent with occupancy across timesteps in a principled probabilistic manner. Therefore, our method estimates velocities from depth measurements without requiring explicit data association across frames!

  \section{Computing depth probabilities using raymarching}
\label{app:max-depth-prob}



\noindent The depth probability of a cell in the grid is the probability that the depth of the scene along the cell's direction is the cell's location. In other words, it is the probability that a visible surface exists in the cell i.e. the cell is occupied and all other occluding cells are empty. Once we compute the depth probability of each cell in the grid, we can place a curtain that lies on the cells with the highest depth probability.

How do we compute the depth probability in a probabilistic occupancy grid? We borrow the idea of ``ray marching'' from the literature on volumetric rendering~\cite{tulsiani2017multi,mildenhall2020nerf}. In order to reconstruct the implicit depth surface from a probabilistic volume, ray marching travels along a ray originating from the sensor and computes the probability of visibility and occlusion at each point. \citet{tulsiani2017multi} performs this for discretized 3D grids (similar to our case) whereas NeRFs~\cite{mildenhall2020nerf} perform this in a continuous space using neural radiance fields. Consider an example of raymarching in Fig.~\ref{fig:dog}c. Let the sequence of cells on a ray be indexed as $1, 2, \dots, n, \dots N$. Recall from \app~\ref{app:dynamic-occupancy-grids} that $\omega^i_t$ is the occupancy probability of the $i$-th cell at timestep $t$. The depth probability of the $n$-th cell $\Pd(n) = \omega^n_t~\prod_{i=1}^{n-1} (1 - \omega^i_t)$ is product of the probabilities that the $n$-th cell is occupied ($\omega^n_t$) and the probabilities that each $i$-th cell on the ray before the $n$-th cell is unoccupied ($1-\omega^i_t$) so that light can reach the $n$-th cell unoccluded. Let us define the ``visibility'' probability $\Pv(n) = \prod_{i=1}^n (1-\omega^i_t)$ that all cells are visible upto the $n$-th cell. Then, we have the following recursive equations:
\begin{align*}
    \Pv(i) &= \Pv(i-1)~(1-\omega^i_t)\\
    \Pd(i) &= \Pv(i-1)~\omega^i_t 
\end{align*}
These recursive equations can be used to compute the depth probability of each cell along a ray efficiently in time $O(N)$ linear in the number of cells on that ray. This strategy is implemented as follows. For each camera ray, we perform the ray-marching procedure and compute the depth probability of each cell along that ray. Then, for each camera ray, we place the curtain at the cell with the maximum depth probability.

  \section{Maximizing information gain}
\label{app:max-info-gain}

\noindent Consider the dynamic Bayes network shown in Fig.~\ref{fig:dbn}b. Given a forecasted prior belief $P(x_t) = \overline{bel}(x_t)$, the information gain framework prescribes that the action $u_t$ should be taken that maximizes the information gain $\IG(x_t, z_t \mid u_t)$ between the state $x_t$ and the observations $z_t$ when taking an action $u_t$. Information gain is a well-studied quantity in information theory and is usually defined as:
\begin{equation}
    \hspace{-0.9em}\IG(x_t, z_t \mid u_t) = \underbrace{\HH(P(x_t)) \vphantom{\big[\big]}}_\text{entropy of $x_t$} - \underbrace{\mathbb{E}_{z_t \mid u_t} \big[ \HH(P(x_t \mid z_t, u_t)) \big]}_\text{conditional entropy of $x_t \mid z_t$ under $u_t$}
    \label{eqn:ig-2h-estimator}
\end{equation}
The information gain is the expected reduction in entropy (i.e. uncertainty) in $x_t$ before and after taking the action $u_t$. \citet{ancha2020eccv} showed that under certain assumptions, the information gain of conventional occupancy grids on placing light curtains is equal to the sum of binary entropies of the occupancy probabilities of the cells that the curtain lies on.

While information gain for conventional occupancy grids is straightforward to derive, it is not so for the case of dynamic occupancy grids. This is because the underlying state space of dynamic occupancy grids is a `mixture' of discrete and continuous spaces. Consider the state $x^i_t$ of the $i$-th cell in the grid. The space of the state $x^i_t$ is:
\begin{equation*}
    x^i_t \in \underbrace{\{\mathrm{unoccupied}\}}_\text{discrete space}~\cup~\underbrace{\{\mathrm{occupied\ with\ } v^i_t \mid v^i_t \in \mathbb{R}^2\}}_\text{continuous space}
\end{equation*}
The cell can either be unoccupied, or be occupied with a continuous velocity. Unfortunately, the entropy of such mixed discrete-continuous spaces is not well-defined~\cite{gao2017estimating}. Therefore the ``2H-estimator'' in Eqn.~\ref{eqn:ig-2h-estimator} (named so because it contains two entropy terms) cannot be used to calculate information gain since the individual terms on the right hand side are not well-defined.

Fortunately, information gain (unlike entropy) is well-defined for most distributions, including discrete-continuous mixtures~\cite{gao2017estimating}. This is possible by using a more general definition of information gain given by the ``Radon–Nikodym'' derivative~\cite{gao2017estimating}:
\begin{equation}
    \IG(x_t, z_t \mid u_t) = \int_{x_t, z_t} \underbrace{\log \frac{d P_{x, z}}{d P_x P_z}}_\text{Radon–Nikodym derivative} d P_{x, z}
    \label{eqn:ig-radon-nikodym}
\end{equation}
The Radon–Nikodym is well-defined for discrete-continuous mixtures~\cite{gao2017estimating}. When the individual entropy terms of Eqn.~\ref{eqn:ig-2h-estimator} (the 2H-estimator) are well-defined, the more general definition of Eqn.~\ref{eqn:ig-radon-nikodym} reduces to Eqn.~\ref{eqn:ig-2h-estimator}. In other words, the two definitions are consistent.

We will now derive the information gain of dynamic occupancy grids using the more general Radon–Nikodym definition. Here, we derive the information gain of a single $i$-th cell. Let $\omega$ be the occupancy probability of the cell. Let the continuous velocity distribution of the cell be denoted by $P(v)$. Assume that we place a curtain on this cell, and we obtain an observation $z^i_t \in \{0, 1\}$ to be a noisy measurement of the cell's occupancy. This is assuming that we are using a depth sensor that can only partially observe occupancy but cannot directly observe velocities. Let $\afp$ and $\afn$ be the false-positive and false-negative rates of the sensor respectively. Then,
\begin{align*}
    &\IG(x^i_t \mid z^i_t)\\
    &= \int_{x, z}dP_{x,z}~\log\frac{d P_{x, z}}{d P_xP_z}\text{\color{blue}\ \ (Radon-Nikodym formulation)}\\
    &= \underbrace{(1-\omega)(1-\afp)\log\frac{(1-\omega)(1-\afp)}{(1-\omega)P(z^i_t=0)}}_\text{\color{red}unoccupied and undetected} +\\
    &\phantom{==}\underbrace{(1-\omega)~\afp\log\frac{(1-\omega)~\afp}{(1-\omega)P(z^i_t=1)}}_\text{\color{red}unoccupied and detected} +\\
    &\phantom{==} \int_v \underbrace{\omega~P(v)dv~(1-\afn)\log\frac{\omega~P(v)dv~(1-\afn)}{\omega~P(v)dv~P(z^i_t=1)}}_\text{\color{red}velocity v and detected} +\\
    &\phantom{==}\int_v \underbrace{\omega~P(v)dv~\afn\log\frac{\omega~P(v)dv~\afn}{\omega~P(v)dv~P(z^i_t=0)}}_\text{\color{red}velocity v and undetected} \\
    &=(1-\omega)(1-\afp)\log\frac{(1-\afp)}{P(z^i_t=0)} + (1-\omega)~\afp\log\frac{\afp}{P(z^i_t=1)} \\
    &\phantom{=} + \omega~(1-\afn)\log\frac{(1-\afn)}{P(z^i_t=1)} + \omega~\afn\log\frac{\afn}{P(z^i_t=0)}\\
    &= -\big[(1-\omega)(1-\afp) + \omega~\afn \big]\log P(z^i_t=0)\\
    &\phantom{==} -\big[(1-\omega)~\afp + \omega(1-\afn) \big]\log P(z^i_t=1)\\
    &\phantom{==} -\omega~\mathrm{H}(\afn)-(1-\omega)~\mathrm{H}(\afp)\\
    &= - P(z^i_t=0)\log P(z^i_t=0) - P(z^i_t=1)\log P(z^i_t=1)\\
    &\phantom{==}-\omega~\mathrm{H}(\afn)-(1-\omega)~\mathrm{H}(\afp)\\
    &= \mathrm{H}(z) -\omega~\mathrm{H}(\afn)-(1-\omega)~\mathrm{H}(\afp)\\
    &= \mathrm{H}(\omega) {\text{\ \ \ \ (assuming that\ } \afp=\afn = 0)}
\end{align*}
\textbf{Assumption 1:} We assume that the sensor is accurate (i.e. the false positive rate $\afp$ and the false negative rate $\afn$ are both close to zero). Then, the information gain of a single cell due to placing a light curtain on that cell is equal to its binary occupancy entropy $\HH_\mathrm{occ}(\omega) = -\omega~\log_2\omega -(1-\omega)~\log_2(1-\omega)$.

\textbf{Assumption 2}: We assume that all cells are independently distributed. Since the information gain of independently distributed random variables is the sum of information gain of individual variables~\cite{CoverThomasBook}, the total information gain is the sum of binary cross entropies $\HH_\mathrm{occ}(\omega^i_t)$ of the cells that the curtain lies on. This is similar to the information gain in \citet{ancha2020eccv}. However, we have been able to prove this mathematically in the more complex case of mixed discrete-continuous distributions.

This theoretical result is also intuitive -- since the depth sensor measurements only provide information about occupancy and not velocity, it is not surprising that the information gain is equal to the total occupancy uncertainty.

  \section{Advantages of light curtains over conventional depth sensors}
\label{app:lc-sensor-benefits}

LiDARs have long range and high accuracy under strong ambient light. However, compared to light curtains, they have poor vertical resolution ($\leq$128 rows), low frame rate (5-20Hz), and are very expensive ($>$\$20K). Table~\ref{table:lc-lidar-comparison} compares LiDARs and light curtains.

\begin{table}[h]
    \centering
    \begin{adjustbox}{width=0.8\columnwidth,center}
    \begin{tabular}{?c?c|c|} 
        \Xhline{0.8pt}
        & {\hspace{1.75em}LiDAR\hspace{1.75em}} & {Light Curtains}\\
        \Xhline{0.8pt}
        Resolution & {\color{red} 128 rows} & {\color{PineGreen} 1280 rows}\\
        \hline
        Cost & {\color{red}$\sim$\$20,000} & {\color{PineGreen}$\sim$\$1,000}\\
        \hline
        Frame rate & {\color{red}10-20 Hz} & {\color{PineGreen}45-60 Hz}\\
        \Xhline{0.8pt}
    \end{tabular}
    \end{adjustbox}
    \caption{Comparison between a modern Ouster OS1~\cite{ousteros1} LiDAR, and Programmable Light Curtains~\cite{bartels2019agile}.}
    \label{table:lc-lidar-comparison}
\end{table}

Passive RGBD sensors (stereo sensors that do not project light) have high spatial and temporal resolution and are inexpensive. However, their accuracy is poor in non-textured regions due to inaccuracies in stereo feature matching.

Active RGBD sensors (like the Kinect sensor that projects light) inherit the benefits of stereo sensors but also work in texture-less regions. However, they have virtually no range outdoors.

Light curtains combine the best of these sensors. They have a long range (nearly 35-50m) both outdoors and indoors, high spatial resolution (1280 rows) and temporal resolution (45-60 Hz), work for textured and texture-less regions, and are inexpensive ($<$1\$K). These advantages have been demonstrated in previous works on programmable light curtains~\cite{bartels2019agile,raaj2021cvpr}.
  \section{Using an extra grid for thread-safety and efficiency}
\label{app:extra-grid}

\noindent Our parallelized pipeline contains three threads: (1) light curtain sensing, (2) Bayes filtering using dynamic occupancy grids, and (3) computing curtain placements. How many grids are required to run these threads in parallel, especially threads 2 and 3?

The motion update step (Eqns.~\ref{eqn:generic-motion-update}, \ref{eqn:dog-motion-update}) moves particles across the grid according to the motion model. The particles cannot be moved in place inside the same grid since it may cause the same particle to be erroneously moved more than once. Therefore, the motion update step requires two grids: a ``source/current" grid and a ``destination/next" grid. Particles from the current grid are copied, moved and placed in the next grid. After the motion update is complete, the roles of the current and next grids are swapped. The next grid is now assigned to be the new ``current" grid since it is now the most up-to-date, incorporating the latest measurements. In the next motion update step, particles move from this grid to the older current grid (now taking the role of the ``next" grid).

The curtain computation thread also performs a motion update when it needs to forecast the current grid to a future timestep (when the next curtain is expected to be imaged). It uses the current grid of the Bayes filtering thread as the source, but requires a third, "forecasting" grid as a destination grid.

Although three grids are sufficient to implement parallelization, the pipeline can be made more efficient. Specifically, consider the situation where two motion updates take place simultaneously from ``current" to ``next" grids in the Bayes filtering thread and ``current" to "forecasting" grid in the curtain computation thread. Once the motion update in the Bayes filtering thread is complete, it cannot immediately perform the next motion update step. It must wait for the curtain computation thread to finish forecasting using the ``current" grid before the ``current" and ``next" grids can be swapped and the next motion update dirties the ``current'' grid. If we use an additional ``extra" grid, the Bayes filtering thread can use this as the destination grid for its next motion update step without needing to wait on the curtain computation thread to finish the latter's forecasting step.

Our parallelized pipeline tightly integrates the three inter-dependent processes in a closed loop. We use a total of four grids to simultaneously guarantee the following two properties: (1) grids in use are never mistakenly overwritten, and (2) no thread ever needs to wait on another to finish processing.

  \begin{figure*}[th!]
    \centering
    \subfloat[HSV colorwheel used to visualize velocities.]{
        \includegraphics[trim=0 0 0 0,clip,width=0.37\textwidth]{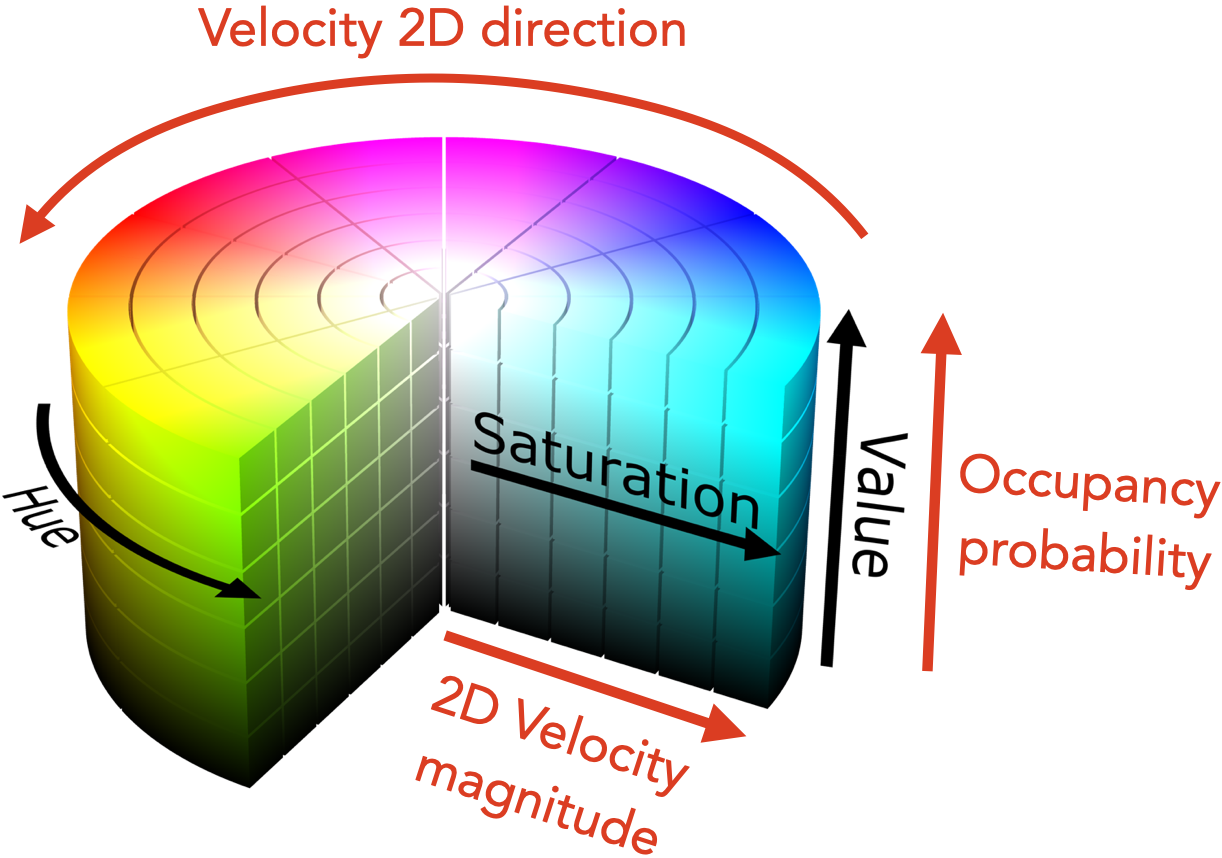}
    }
    \subfloat[Colorwheel from the top-down view]{
        \hspace{0.5cm}
        \includegraphics[trim=0 -3cm 0 0,clip,width=0.15\textwidth]{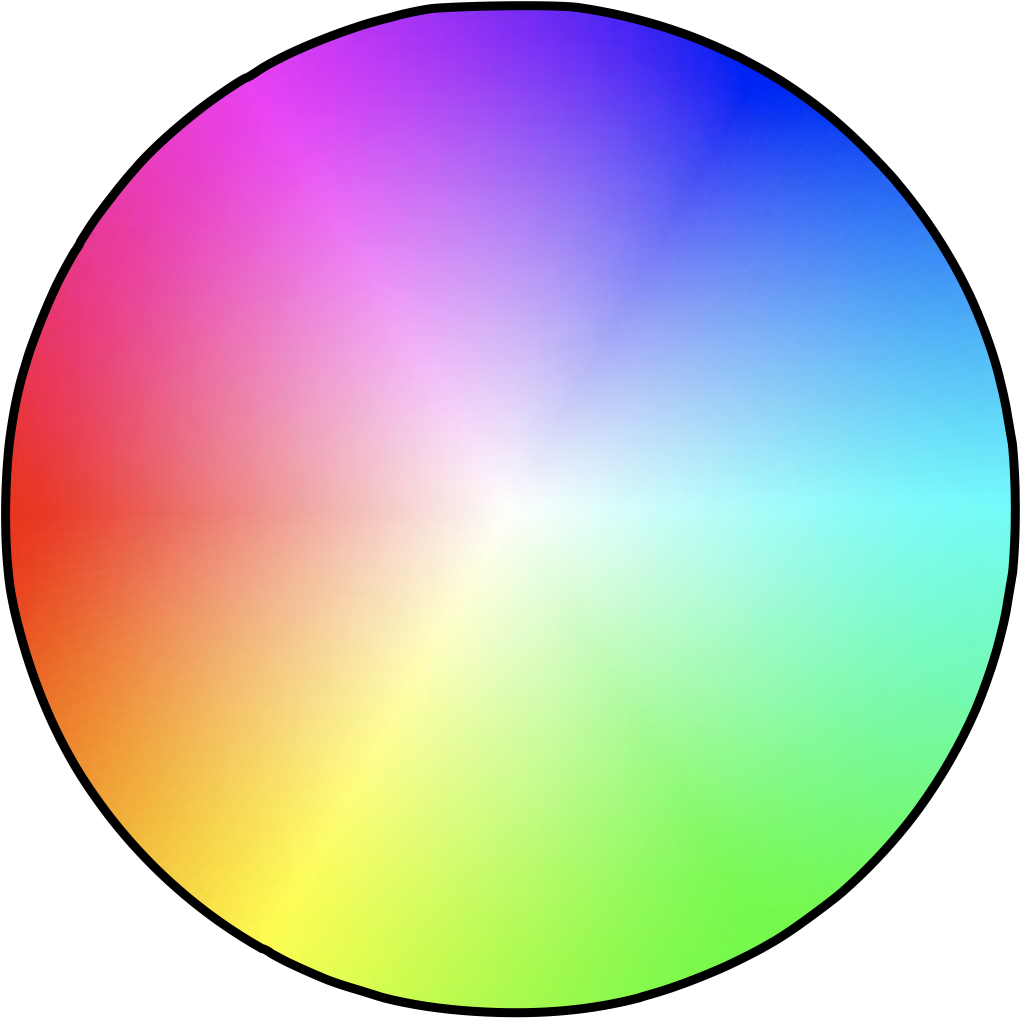}
        \hspace{0.75cm}
    }
    \subfloat[GT velocities of the simulated environment.]{
        \includegraphics[trim=0 0 120 0,clip,width=0.35\textwidth]{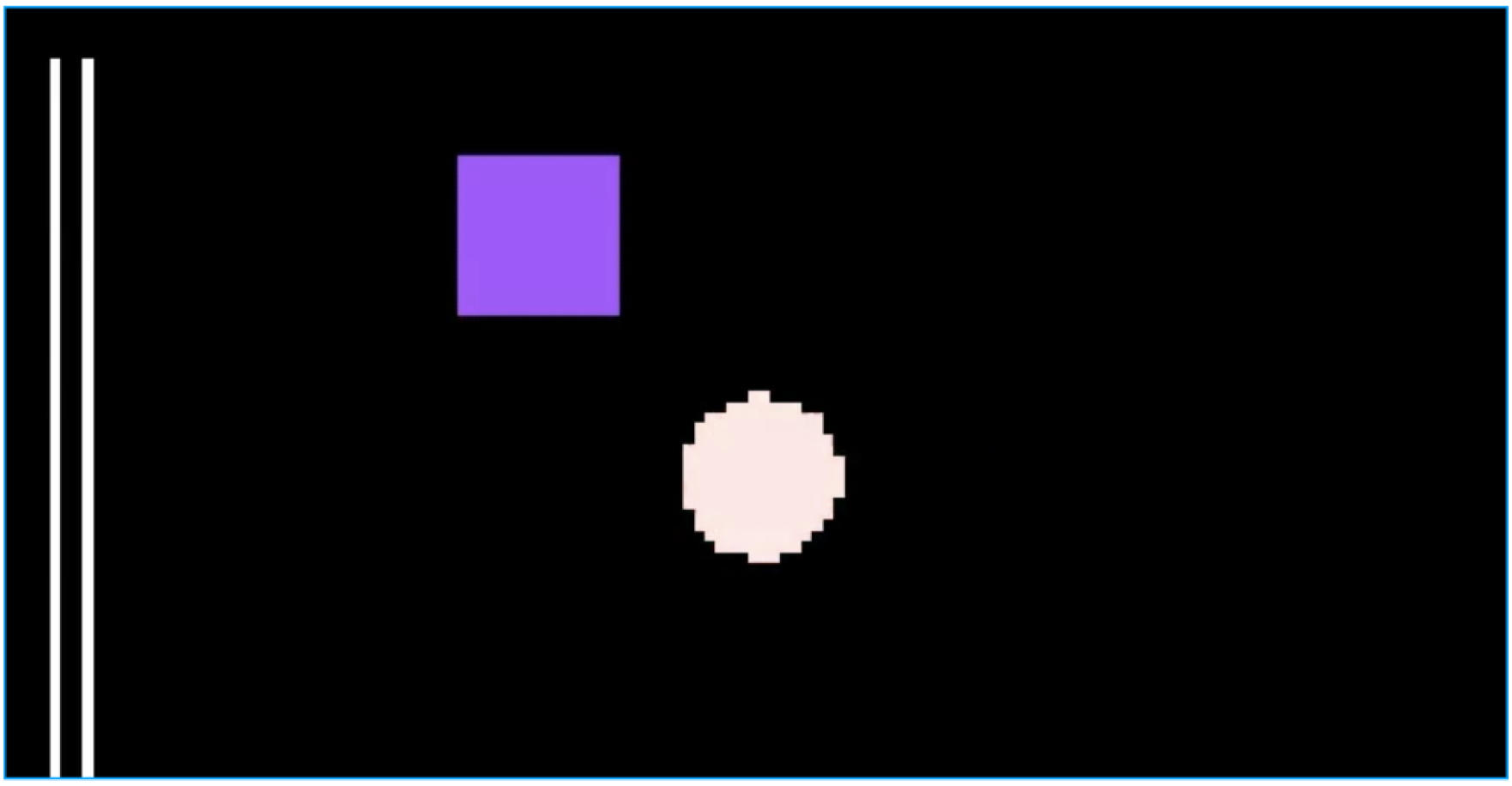}
    }
    \caption{(a) The HSV (hue-saturation-value) colorwheel~\cite{HSV} used to visualize 2D velocities and occupancies. Value corresponds to the occupancy probability. The hue corresponds to the direction of the velocity. Saturation corresponds the magnitude of velocity. (b) The HSV colorwheel in the top-down view, denoting the direction of velocity. (c) Ground truth velocities and occupancies in the simulated environment. The grid shows a stationary wall to the left and two objects. The bluish-purple square is moving upwards in 2D (i.e. farther away from the sensor in 3D) whereas the other objects in white are stationary.}
    \label{fig:hsv-colorwheel}
\end{figure*}

  \section{Evaluation Metrics}
\label{app:evaluation-metric}

\newcommand{\pr}[1]{\widetilde{o}^{\hspace{0.2em}#1}_{t+\Delta t}}
\newcommand{\gt}[1]{o^{#1}_{t+\Delta t}}
\newcommand{\ilos}{\mathcal{I}_\mathrm{LOS}}

As mentioned in Sec.~\ref{sec:evaluation-metric}, \textit{forecasted occupancy}~\cite{mahjourian2022occupancy,waymo2022challenge} simultaneously captures both the accuracy of occupancy estimates as well as velocity estimates. This metric is especially pertinent for obstacle avoidance where future occupancy of obstacles is needed to plan paths that avoid collisions.

We will use the notation for dynamic occupancy grids introduced in Sec.~\ref{app:mathematical-framework}. The current dynamic occupancy grid at timestep $t$ is represented by $G_t = \{\omega^i_t, V^i_t \mid i \in \mathcal{I}\}$, where $\omega^i_t$ is the Bernoulli occupancy probability of the $i$-th cell, and $V^i_t = \{ (v^{i, m}_t, p^{i, m}_t) \mid 1 \leq m \leq M\}$ is the set of $M$ weighted particles that represents the velocity distribution of the $i$-th cell.

To evaluate $G_t$, we first apply the motion update step (Eqns.~\ref{eqn:generic-motion-update}, \ref{eqn:dog-motion-update}) to forecast it by a time $\Delta t$ and obtain the dynamic occupancy grid $G_{t+\Delta t}$ at time $t + \Delta t$. Then, the forecasted occupancy probabilities $\{\omega^i_{t + \Delta t} \mid i \in \mathcal{I}\}$ are evaluated against the ground truth occupancies $\{\gt{i} \mid i \in \mathcal{I}\}$. We follow prior works~\cite{hornung2013octomap,meyer2012occupancy,tatarchenko2017octree} that treat the evaluation of occupancy as a classification problem and compute binary occupancies $\pr{i} = \mathbb{I}(\omega^i_{t + \Delta t} \geq 0.5)$ thresholded at 0.5 probability.

We use the following metrics to evaluate the quality of predicted occupancy. We ignore cells that are occluded (in the ground truth) since (1) they cannot be observed by optical sensors, and (2) they are not the closest object to the robot making them less relevant for obstacle avoidance. Let $\ilos$ be the subset of cells that are in the sensor's line-of-sight (LOS). Note that $\pr{i}, \gt{i} \in \{0, 1\}$.

\begin{enumerate}[leftmargin=*]
    \item \textit{Classification accuracy:} The fraction of cells whose occupancy is correctly predicted.
    \begin{equation*}
        \texttt{Accuracy} = \frac{\sum_{i \in \ilos} \mathbb{I}\{\pr{i} = \gt{i}\}}{|\ilos|}
    \end{equation*}
    \item \textit{Precision:} The fraction of cells predicted to be occupied that were actually occupied.
    \begin{equation*}
        \texttt{Precision} = \frac{\sum_{i \in \ilos} \mathbb{I}\{\pr{i} = 1\}~\mathbb{I}\{\gt{i} = 1\}}{\sum_{i \in \ilos} \mathbb{I}\{\pr{i} = 1\}}
    \end{equation*}
    \item \textit{Recall:} The fraction of occupied cells that were also predicted to be occupied.
    \begin{equation*}
        \texttt{Recall} = \frac{\sum_{i \in \ilos} \mathbb{I}\{\pr{i} = 1\}~\mathbb{I}\{\gt{i} = 1\}}{\sum_{i \in \ilos} \mathbb{I}\{\gt{i} = 1\}}
    \end{equation*}
    \item \textit{\fone-Score:} A combination (harmonic mean) of precision and recall that is a commonly used for binary classification~\cite{f1Score}. The \fone~score is robust to class imbalance; unlike precision and recall, it cannot be trivially improved by predicting mostly negative labels and mostly positive labels.
    \begin{equation*}
        \texttt{\fone-Score} = \frac{2 \cdot \texttt{Precision} \cdot \texttt{Recall}}{\texttt{Precision} + \texttt{Recall}}
    \end{equation*}
    \item \textit{IoU:} The intersection-over-union between cells that are occupied (in ground truth) and cells that are predicted to be occupied.
    \begin{equation*}
        \texttt{IoU} = \frac{\sum_{i \in \ilos} \mathbb{I}(\pr{i} = 1)~\mathbb{I}(\gt{i} = 1) }{\sum_{i \in \ilos} \mathbb{I}(\pr{i} = 1 \text{\ or\ } \gt{i}=1)}
    \end{equation*}
\end{enumerate}
For all metrics, a higher numerical score is better.
  \section{Visualizing velocities and occupancies}
\label{app:colorwheel}

\noindent Fig.~\ref{fig:hsv-colorwheel} shows how we visualize 2D velocities and occupancies (both ground truth and estimated velocities and occupancies). The visualization of an example ground truth grid is shown in Fig.~\ref{fig:hsv-colorwheel}c. We use the three-dimensional HSV colorwheel shown in Fig.~\ref{fig:hsv-colorwheel}a to jointly visualize velocities (two-dimensional) and occupancies (one-dimensional). The `value' encodes the occupancy probability; dark means low occupancy probability and bright means high occupancy probability. The `hue' encodes the direction of velocity. We show a top-down view of the HSV colorwheel in Fig.~\ref{fig:hsv-colorwheel}b for clarity. For example, the bluish-purple hue of the cuboid in Fig.~\ref{fig:hsv-colorwheel}c means that the cuboid is moving upwards in 2D i.e. away from the sensor in 3D. `Saturation' encodes the magnitude of velocity. This means that white is stationary (e.g. the walls of the environment shown as parallel white lines) whereas colorful regions corresponds to high speed.

  \section{Non-stationary rewards}
\label{app:non-stationary-rewards} 

\noindent Vanilla multi-armed bandits assume that the reward distribution for each action is stationary. The Q-value of an action after it was been performed $n$ times is computed as $Q_n = \frac{1}{n} \sum_{i=1}^n R_i$, where $R_i$ is the reward obtained in the $i$-th trial. This is equivalent to the following recursive update rule: $Q_{n+1} = Q_n + \frac{1}{n}[R_n - Q_n]$, where the Q-value is incremented by the error scaled by a decaying factor $\frac{1}{n}$.

However, in our case, a single placement strategy may not be superior to the rest at all times and in all situations. The reward distribution for each strategy (action) may change with time. Hence, we assume that our rewards are non-stationary. For non-stationary rewards, we wish to give more weight to recent rewards than to older rewards. Therefore, the decaying parameter is replaced by a constant step-size parameter $\alpha$: $Q_{n+1} = Q_n + \alpha~[R_n - Q_n]$. This weights newer rewards exponentially more than older rewards according to the expression: $Q_n = (1-\alpha)^{n-1}~R_1 + \sum_{i=2}^n \alpha~(1-\alpha)^{n-i}~R_i$.

  \section{Efficient Light Curtain Simulation}
\label{app:efficient-simulation}

\definecolor{deeporange}{rgb}{0.8627451 , 0.34509804, 0.16470588}
\newcommand{\tikzcircle}[2][ForestGreen,fill=ForestGreen]{\tikz[baseline=-0.5ex]\draw[#1,radius=#2] (0,0) circle ;}%

\begin{figure}[t!]
    \centering
    \includegraphics[trim=0 0 0 0,clip,width=0.44\textwidth]{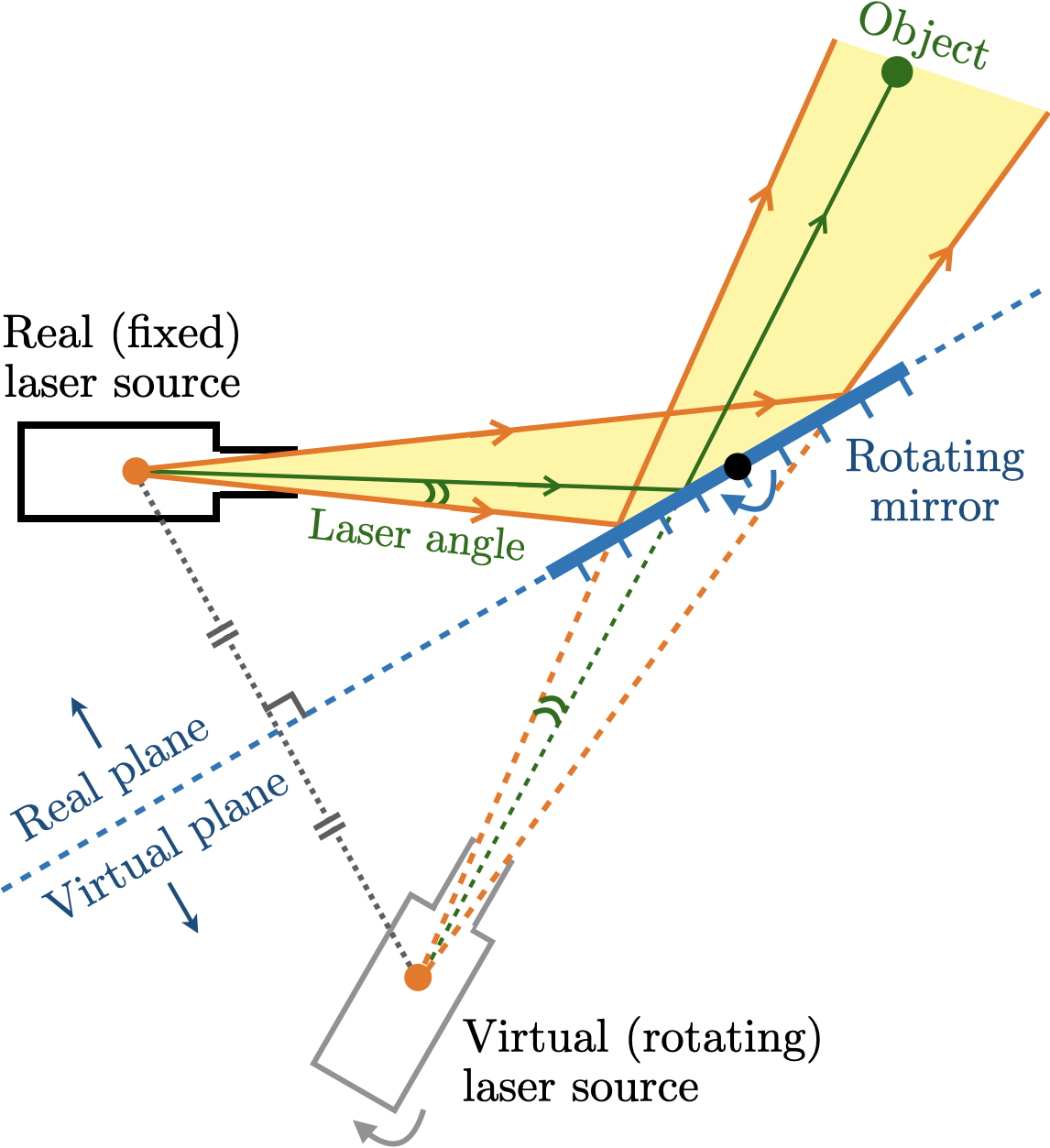}
    \caption{\textit{Efficient light curtain simulation using a virtual laser.} A light curtain consists of a fixed laser source that emits a light beam, and a rotating galvo-mirror that reflects and redirects the light in any desired direction. The laser beam is collimated to a thin rectangular sheet; however, in reality it is a prismatic slab containing a small divergence. The pixel intensity of an object (shown by a {\color{ForestGreen}green circle \tikzcircle{0.1}}) imaged by the light curtain depends on the radiant intensity of the laser ray that is incident on the object (shown by the {\color{ForestGreen}green ray $\rightarrow$}). A ray at the center of the beam has the highest intensity whereas a ray at the boundary of the beam (shown by {\color{deeporange}orange rays $\rightarrow$}) has the lowest intensity. In order to simulate light curtain pixel intensities for a given object, we must compute its incident ray i.e. compute its ``laser angle" shown by \includegraphics[height=0.8em]{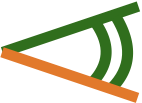}. This computation can be expensive since (1) it involves tracing rays between the source and the object through a reflection at the mirror, and (2) it must be performed for each pixel. Our insight is to construct a ``virtual'' laser source by reflecting the real source about the mirror plane. Due to the laws of reflection, the reflected beam is equivalent to originating from the virtual source behind the mirror. This allows the laser angle to be computed efficiently by projecting the object point in the virtual source's frame. Furthermore, the virtual source needs to be reflected only once for each mirror configuration; all pixel points in the currently active camera column can be efficiently projected into the same virtual laser source.}
    \label{fig:efficient-simulation}
\end{figure}

Fig.~\ref{fig:efficient-simulation} shows the working principle behind the illumination module of a programmable light curtain.
It consists of a fixed laser source that emits a light beam, and a rotating galvo-mirror that reflects and redirects the light in any desired direction. The laser beam is collimated to a thin rectangular sheet; however, in reality it is a prismatic slab containing a small divergence. The pixel intensity of an object (shown by a {\color{ForestGreen}green circle \tikzcircle{0.1}}) imaged by the light curtain depends on the radiant intensity of the laser ray that is incident on the object (shown by the {\color{ForestGreen}green ray $\rightarrow$}). A ray at the center of the beam has the highest intensity whereas a ray at the boundary of the beam (shown by {\color{deeporange}orange rays $\rightarrow$}) has the lowest intensity. In order to simulate light curtain pixel intensities for a given object, we must compute its incident ray i.e. compute its ``laser angle" shown by \includegraphics[height=0.8em]{figures/angle-icon.png}.

Computing the laser ray incident to the object is expensive because (1) it involves tracing rays between the source and the object through a reflection at the mirror, and (2) it must be performed for each pixel. Our insight is to construct a ``virtual'' laser source by reflecting the real source about the mirror plane. Due to the laws of reflection, the reflected beam is equivalent to originating from the virtual source behind the mirror. This allows an object point's laser angle to be computed efficiently by projecting it in the virtual source's frame. Furthermore, the virtual source needs to be reflected only once for each mirror configuration; all pixel points in the currently active camera column can be efficiently projected into the same virtual source.
  \section{Full-stack navigation}
\label{app:full-stack-navigation}

\begin{figure*}[t!]
    \centering
    \hspace*{-1.6cm}
    \includegraphics[trim=0 0 0 0,clip,width=1.17\textwidth]{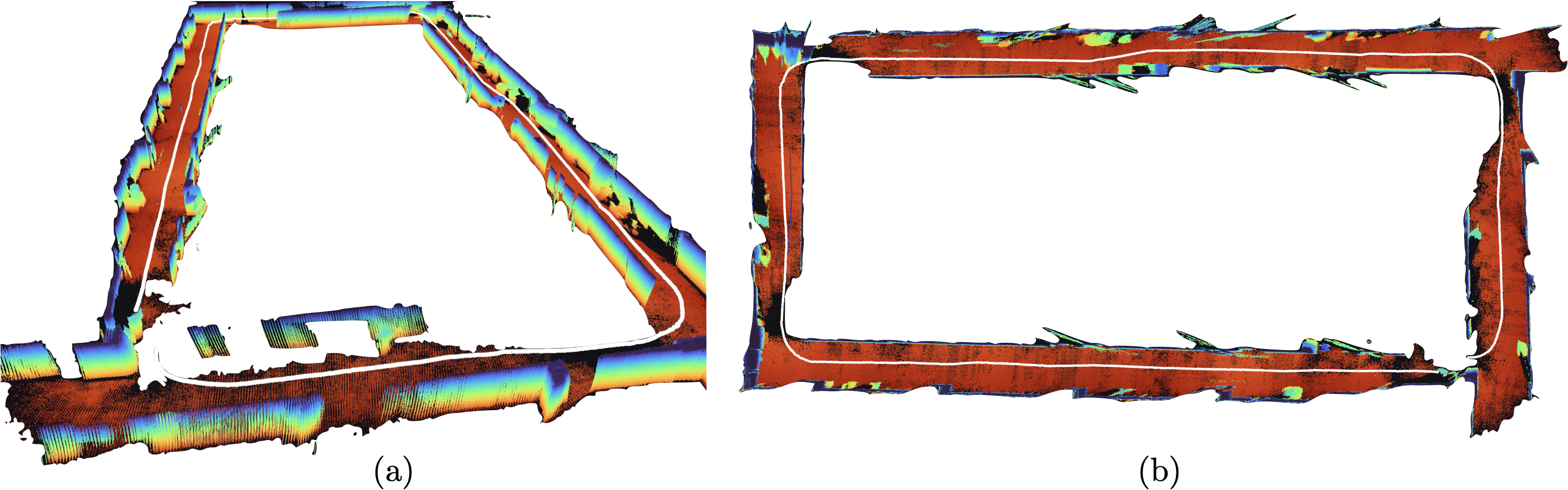}
    \caption{\textit{Dense indoor reconstruction and mapping using our integrated system.} The light curtain was mounted on a mobile robot (Fig.~\ref{fig:wheelchair-envs}a) and operated in an indoor hallway environment. Detected depth points from light curtains were input to ORB-SLAM3~\cite{campos2021orb}, an RGB-D based localization and mapping system that estimates the robot's pose. The pose estimates are fed back into our pipeline to perform ego-motion subtraction in the motion update step. The robot trajectory is shown as a white line. (a) \textit{Sideways perspective view:} showing dense reconstruction of walls, the floor and other objects. (b) \textit{Top-down orthographic view:} showing the accuracy of localization (white line) and loop-closure. Please find the full video on the \href{\website}{project website}. This experiment serves to demonstrate our full-stack navigation pipeline using light curtains.}
    \label{fig:nsh-reconstruction}
\end{figure*}

\begin{figure*}[t!]
    \centering
    \includegraphics[trim=0 0 0 0,clip,width=0.75\textwidth]{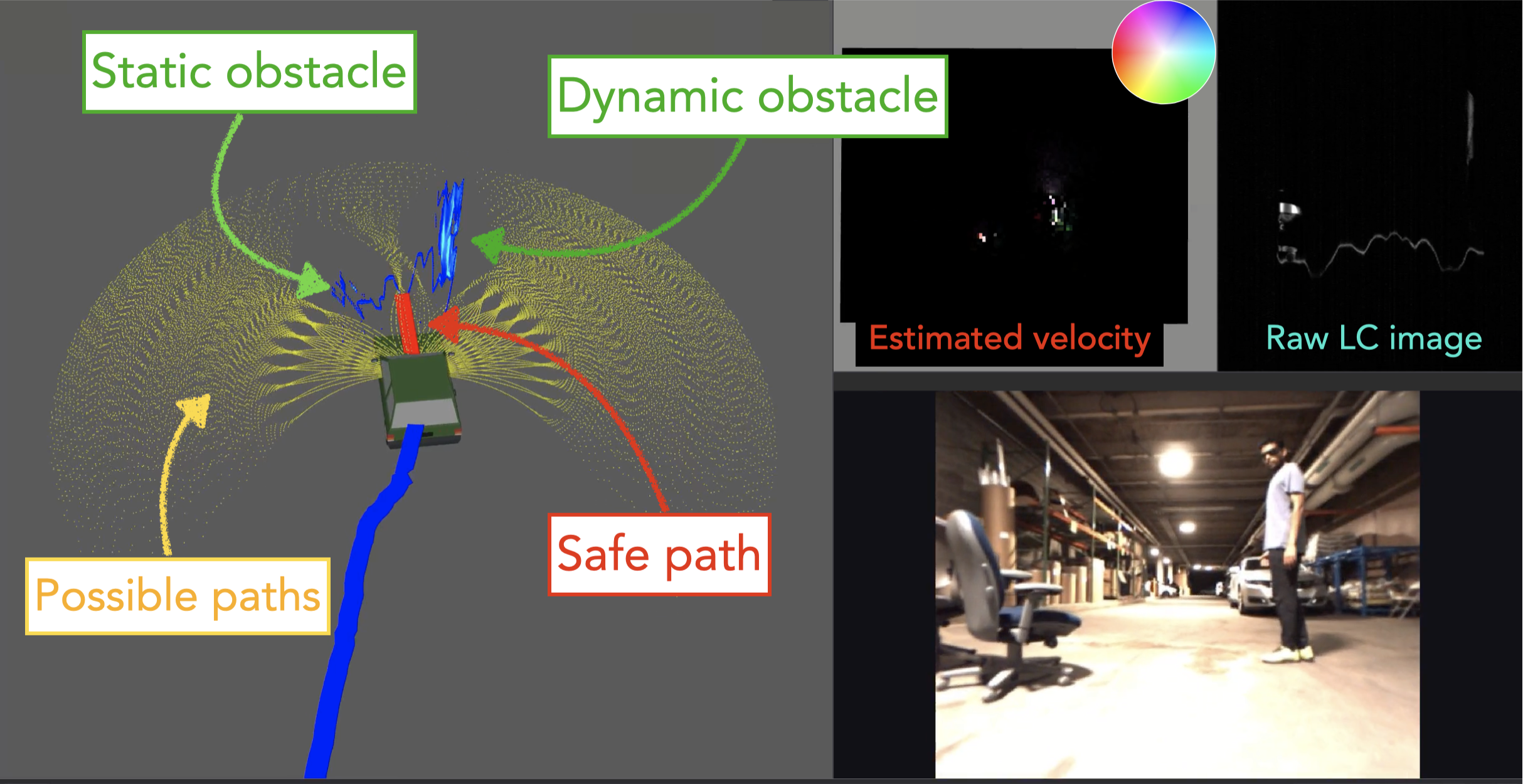}
    \caption{\textit{Real-time obstacle avoidance using our integrated system.} The robot is represented by a green vehicle. The yellow curves show a library of dynamically feasible paths of the robot~\cite{cao2021tare}. Points detected by the light curtain on the static (chair) and dynamic (person) objects are show in blue. The feasible paths that are expected to collide with objects are removed, and a safe path (shown in red) is chosen by the planner.  Please find the full video on the \href{\website}{project website}.}
    \label{fig:obstacle-avoidance}
\end{figure*}

We integrate our system into a full-stack navigation pipeline based on the Autonomous Exploration Development Environment~\cite{cao2022autonomous}. We mount the light curtain device on a mobile robot (see Fig.~\ref{fig:wheelchair-envs}a). Our tightly integrated pipeline performs localization, mapping, planning, control and obstacle avoidance. We use ORB-SLAM3~\cite{campos2021orb} for localization and mapping that takes depth from light curtains as input while the planning and control capabilities are provided by \citet{cao2022autonomous}. The pipeline described in Sec.~\ref{sec:parallelized-pipeline} combines  light curtain placement strategies using self-supervised multi-armed bandits with recursive Bayes estimation of dynamic occupancy grids. The output of this pipeline i.e. position and velocity estimates are used by the autonomy stack to perform dense mapping in an indoor environment and obstacle avoidance. Furthermore, the localization from ORB-SLAM3~\cite{campos2021orb} is fed back into our pipeline for ego-motion subtraction in the motion update step (Eqns.~\ref{eqn:generic-motion-update}, \ref{eqn:dog-motion-update}). We show two demonstrations of our fully integrated autonomy stack:

\subsection{Real-time dense mapping}

Light curtains sense objects that intersect its surface at a high resolution. This ability can be leveraged to perform dense mapping and reconstruction of an environment. Please see a video of dense real-time reconstruction of an indoor hallway environment using our system on the \href{\website}{project website}. Fig.~\ref{fig:nsh-reconstruction} shows a sideways and top-down projection of the same. The robot was operated in the indoor hallway environment, and 3D points detected by light curtains were input to ORB-SLAM3~\cite{campos2021orb}, an RGB-D based localization and mapping system that estimates the robot's pose. The pose estimates are fed back into our pipeline to perform ego-motion subtraction in the motion update step. The robot trajectory is shown as a white line. Fig.~\ref{fig:nsh-reconstruction}a shows that the floor, walls and other objects are reconstructed densely and accurately. Fig.~\ref{fig:nsh-reconstruction}b contains a top-down orthographic view which shows the accuracy of our system's localization -- the robot's trajectory was correctly determined to be an (approximately) closed loop around the building floor.
This experiment serves to demonstrate our full-stack navigation pipeline using light curtains.

\subsection{Real-time obstacle avoidance}

Light curtains are a fast sensor ($\sim$45 Hz) whose speed is leveraged by our system to produce position and velocity estimates at a high frequency ($\sim$35 Hz). We use these estimates for real-time obstacle avoidance, shown in Fig.~\ref{fig:obstacle-avoidance}. The robot is represented by a green vehicle. The yellow curves show a library of dynamically feasible paths of the robot~\cite{cao2021tare}. Points detected by the light curtain are shown in blue. There are two objects in the scene: a static chair, and a moving person. Using the position estimates of obstacles, feasible paths that are expected to collide with objects are rejected, and a safe path (shown in red) is chosen by a local planner~\cite{cao2021tare}. Please find the full video of real-time obstacle avoidance on the \href{\website}{project website}.

\end{appendices}

\end{document}